\documentclass[journal]{IEEEtran}
\usepackage{url}

\usepackage{wrapfig,booktabs,amsmath,colortbl} 
\usepackage{amsfonts,amssymb,bbding,mathtools,amsthm,pifont}
\usepackage{graphicx}
\usepackage{subfigure}
\usepackage{xspace}
\usepackage{xcolor}
\definecolor{pink}{HTML}{747199}

\usepackage[colorlinks=true,citecolor=pink,urlcolor=pink,linkcolor=pink]{hyperref} 
\usepackage{threeparttable}
\usepackage{adjustbox}

\usepackage{enumitem}
\usepackage[numbers,compress,sort&compress]{natbib}
\usepackage{microtype}     
\usepackage{chngcntr}      
\usepackage{multirow}
\usepackage{tikz}
\usepackage{amssymb} 
\usetikzlibrary{trees,positioning,shapes,shadows,arrows.meta}
\usepackage[edges]{forest}
\usepackage[capitalize,noabbrev]{cleveref}
\usepackage{caption, subcaption}
\usepackage[textsize=tiny]{todonotes}
\definecolor{yellow}{HTML}{cda380}

\usepackage{algorithmic}
\usepackage{algorithm}
\definecolor{lightgray}{gray}{0.94}
\definecolor{deepgray}{gray}{0.5}
\newcommand{\true}{\Checkmark}
\newcommand{\false}{-}
\theoremstyle{plain}
\newtheorem{theorem}{Theorem}[section]

\theoremstyle{definition}
\newtheorem{definition}[theorem]{Definition}

\theoremstyle{remark}

\newcommand{\T}{\mathrm{T}}

\newcommand{\ie}{\textit{i}.\textit{e}., }
\newcommand{\eg}{\textit{e}.\textit{g}., }

\title{Deep Time-Series Forecasting in 10 Years: A Survey}
\author{Hao Wang, Licheng Pan, Qingsong Wen, Jialin Yu, Zhichao Chen, Chunyuan Zheng,\\Xiaoxi Li, Zhixuan Chu, Chao Xu, Mingming Gong, Haoxuan Li,  Yuan Lu,\\ Zhouchen Lin~\IEEEmembership{Fellow,~IEEE}, Philip Torr, Yan Liu~\IEEEmembership{Fellow,~IEEE}
\thanks{Zhouchen Lin was supported by the Beijing Major Science and Technology Project (No. Z251100008425006), the NSF China (No. 62276004), and the Beijing Natural Science Foundation (No. L257007).
Haoxuan Li was supported by the Shanghai General AI Foundation Models Program.
Zhichao Chen was supported by the NSF China (No. 62606018).
Jialin Yu acknowledges support from Microsoft Ltd and this work is partially supported by the UKRI Turing AI Fellowship (EP/W002981/1).}
\thanks{Hao Wang, Zhichao Chen, and Zhouchen Lin are with the School of Intelligence Science and Technology, Peking University, Beijing 100871, China (e-mail: Ho-ward@outlook.com; 2506399003@pku.edu.cn; zlin@pku.edu.cn).}
\thanks{Licheng Pan and Zhixuan Chu are with the School of Cyber Science and Technology, Zhejiang University, Hangzhou, Zhejiang 310027, China (e-mail: 12521034@zju.edu.cn; zhixuanchu@zju.edu.cn).}
\thanks{Qingsong Wen is with Squirrel AI Learning, Bellevue, WA 98004, USA (e-mail: qingsongedu@gmail.com).}
\thanks{Jialin Yu and Philip Torr are with The Department of Engineering Science, University of Oxford, Oxford, UK (e-mail: yu.jialin@outlook.com; philip.torr@eng.ox.ac.uk).}
\thanks{Chunyuan Zheng is with the School of Mathematical Sciences, Peking University, Beijing 100871, China (e-mail: cyzheng@stu.pku.edu.cn).}
\thanks{Xiaoxi Li and Yuan Lu are with Xiaohongshu Incorporation, Shanghai 200025, China (e-mail: xiaoxi\_li@ruc.edu.cn; luyuan3@xiaohongshu.com).}
\thanks{Chao Xu is with the College of Control Science and Engineering, Zhejiang University, Hangzhou, Zhejiang 310027, China (e-mail: cxu@zju.edu.cn).}
\thanks{Mingming Gong is with the School of Mathematics and Statistics, University of Melbourne, Melbourne VIC 3010, Australia (e-mail: mingming.gong@unimelb.edu.au).}
\thanks{Haoxuan Li is with the Institute for Artificial Intelligence, Peking University, Beijing 100871, China (e-mail: hxli@pku.edu.cn).}
\thanks{Yan Liu is with the Department of Computer Science, University of Southern California, Los Angeles, CA 90007, USA  (e-mail: yanliu.cs@usc.edu).}
}

\begin{document}
\maketitle

\begin{abstract}
Autocorrelation is a common property of time-series, where each observation is dependent on its predecessors.
In deep time-series forecasting, it raises two central challenges: \textit{\ding{182} designing backbone architectures to model autocorrelation in history sequences}, and \textit{\ding{183} devising loss functions to model autocorrelation in label sequences}.
Recent studies have made strides in tackling these challenges, but a systematic survey examining both aspects remains lacking.
To bridge this gap, this paper reviews deep time-series forecasting from an autocorrelation modeling perspective, offering two contributions beyond existing surveys.
First, it introduces a taxonomy that jointly covers both backbone architectures and loss functions, whereas prior surveys provide limited coverage of the latter.
Second, it analyzes the motivations and insights underlying the surveyed literature from a unified autocorrelation perspective, providing a holistic overview of the field's evolution.
Additional resources and details are available at \url{https://github.com/Master-PLC/Awesome-TSF-Papers}.
\end{abstract}
 
 
\section{Introduction}

Time-series forecasting aims to predict future values from historical observations. It has demonstrated substantial value across diverse domains~\cite{acmsurvey,tsinghuasurvey}, such as asset pricing and risk management in finance~\citep{mars}, traffic optimization and route planning in intelligent transportation~\citep{application_traffic,application_traffic}, load balancing and grid stability in energy systems~\citep{grabner2023global}, and process monitoring and inferential sensing in manufacturing~\citep{chen2026blending,pan2025tmoe,wang2026taideepfilter,qian2023novel}.
With the rapid advancement of neural networks, deep time-series forecasting has emerged as a prominent paradigm~\cite{qiutfb,tsinghuasurvey}, whose scalability and nonlinear representational capacity enable the capture of complex temporal dependencies in large-scale datasets, yielding substantial gains in forecasting accuracy. Currently, deep time-series forecasting has become a foundational technique for handling the increasingly large, nonlinear, and complex time-series data in modern applications~\cite{wen2022robust,chen2024ocndplvm}.

Autocorrelation is a fundamental property of time-series~\cite{anderson2011statistical}, where each observation is dependent on its predecessors.
For instance, daily temperatures correlate with those of the previous day.
This property distinguishes time-series from generic tabular data, and introduces unique challenges for forecasting tasks~\cite{wang2025iclrfredf}.
The significance of autocorrelation has been recognized since the early days of statistical forecasting, motivating methods that explicitly model it, such as ARIMA and VAR~\citep{VAR, Arima}.
In the era of deep time-series forecasting, autocorrelation remains critical, as it is inherent to both the input (history) and label (future) sequences used to train deep neural networks.
Formally, this raises two research challenges: \textbf{\ding{182} \textit{how to devise backbone architectures to accommodate autocorrelation in the history sequence}}, and \textbf{\ding{183} \textit{how to devise loss functions to accommodate autocorrelation in the label sequence}.} 

Addressing these two challenges has driven recent advances in deep time-series forecasting.
To model history autocorrelation (challenge \ding{182}), diverse backbone architectures have been developed, such as recurrent neural networks (RNNs)~\cite{LSTM,GRU}, convolutional neural networks (CNNs)~\cite{tcn,Filternet,Moderntcn}, dense neural networks (DNNs)~\cite{DLinear,FreTS}, and Transformers~\cite{PatchTST,itransformer,qiudag}.
Recent advances within these architectural families include deep state space models~\cite{gu2023mamba,smamba}, large-kernel CNNs~\cite{Moderntcn,ding2024unireplknet}, deep decomposition models~\cite{wang2024timemixer,wang2025timemixer++}, and large language models (LLMs)~\cite{Aurora}.
These advances overcome limitations of earlier architectures and better model complex autocorrelation patterns.
To model label autocorrelation (challenge \ding{183}), various loss functions have been proposed, grounded in likelihood estimation~\cite{wang2025iclrfredf,wang2026iclrqdf}, shape alignment~\cite{soft-dtw}, and distribution balancing~\cite{wang2026iclrdistdf,AST}.
More recently, a new paradigm reframes forecasting as a conditional generation task~\cite{TimeDiff}, enabling label autocorrelation to be modeled through diffusive~\cite{Diffusion-TS,CNDiff} or autoregressive approaches~\cite{Autotimes}.

Despite rapid progress in backbone architectures and loss functions, a survey that examines both aspects from a unified view remains absent.
First, existing surveys primarily focus on backbone architectures~\citep{wen2023transformersurvey,tsinghuasurvey}. In contrast, loss functions have received comparatively limited attention despite a growing body of work~\cite{wang2025iclrfredf,wang2025nipstimeo1}.
Second, existing surveys often emphasize the technical designs of existing methods, without sufficiently articulating the underlying challenges they address.
In particular, the challenges raised by autocorrelation remain insufficiently analyzed.
This gap highlights the need for a unified framework to understand the core challenges in this field and the evolution of solutions addressing them.

To bridge this gap, this paper reviews deep time-series forecasting from a unified perspective of autocorrelation modeling, which makes two primary contributions.
First, it introduces a taxonomy that comprehensively covers both backbone architectures and loss functions.
On backbone architectures, it covers not only canonical approaches but also emerging ones: state space models~\cite{gu2023mamba,smamba}, mixture-of-experts models~\cite{Time-MoE}, large-kernel CNNs~\cite{Moderntcn}, and LLM-based models~\cite{LLM4TS}, to help readers navigate fast-evolving trends. On loss functions, it provides the first systematic review in this domain, addressing a critical gap left by prior surveys~\cite{wen2023transformersurvey,tsinghuasurvey}.
Second, it analyzes the motivations and insights underlying the surveyed literature from a unified autocorrelation perspective.
Collectively, these contributions offer a unified and holistic framework for understanding recent developments in deep time-series forecasting.

The remainder of this paper is organized as follows.
Section~\ref{sec:preliminaries} formalizes the time-series forecasting problem and highlights the central role of autocorrelation.
Section~\ref{sec:taxonomy} presents the proposed taxonomy and compares it with those in existing surveys.
Section \ref{sec:architecture} reviews backbone architectures designed for modeling history autocorrelation, while Section \ref{sec:loss} reviews loss functions for modeling label autocorrelation.
Section \ref{sec:futureworks} discusses future research directions inspired by existing studies to further exploit autocorrelation.
Finally, the conclusions and limitations of this work are discussed.

\begin{figure}
    \centering
    {\includegraphics[width=0.48\linewidth]{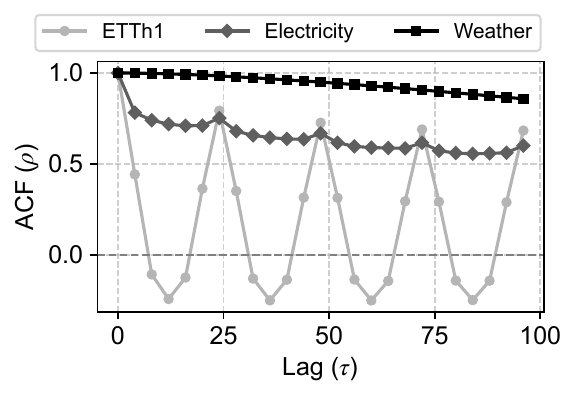}}\hfill
    {\includegraphics[width=0.48\linewidth]{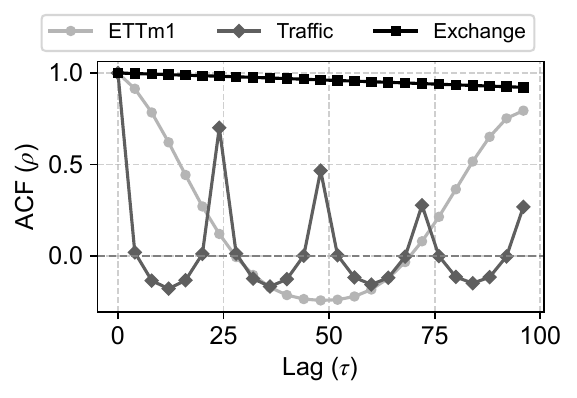}}\hfill
    \caption{The ACF at different lags on time-series datasets.}
    \label{fig:case2}
\end{figure}

\begin{figure*}
    \centering
    \subfigure[Time-series of white noise and its ACF.]{\includegraphics[width=0.48\textwidth]{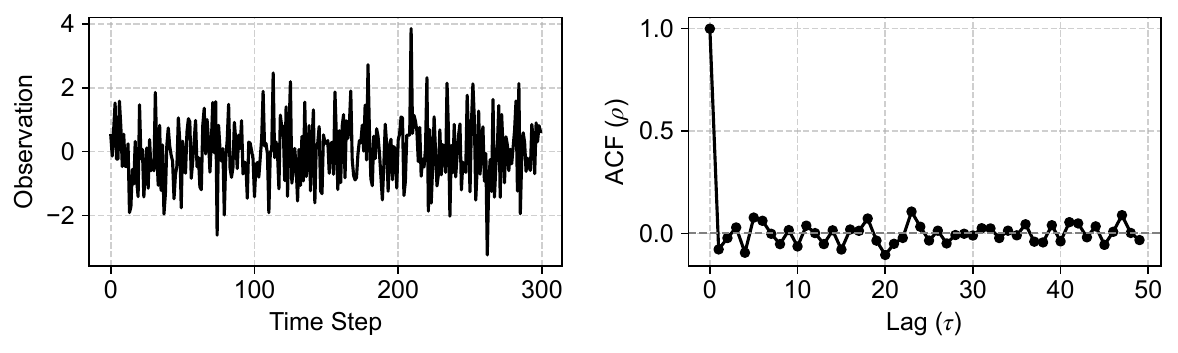}}\hfill
    \subfigure[Time-series with trend pattern and its ACF.]{\includegraphics[width=0.48\textwidth]{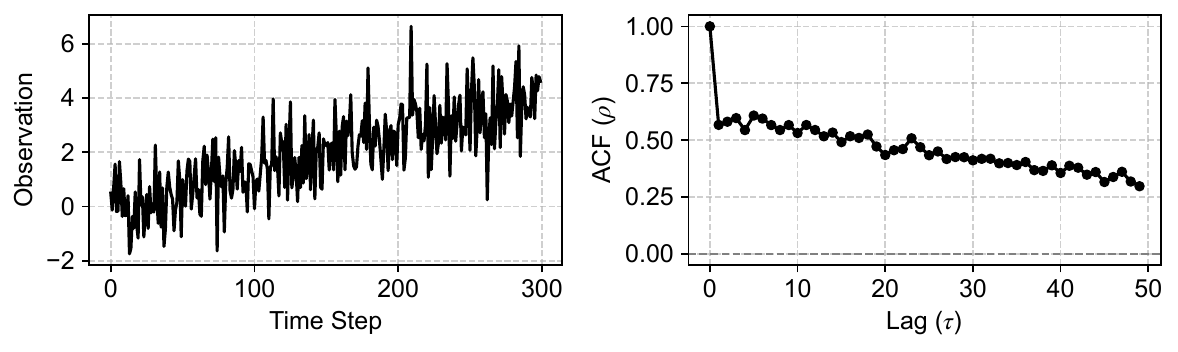}}
    \subfigure[Time-series with cyclic pattern and its ACF.]{\includegraphics[width=0.48\textwidth]{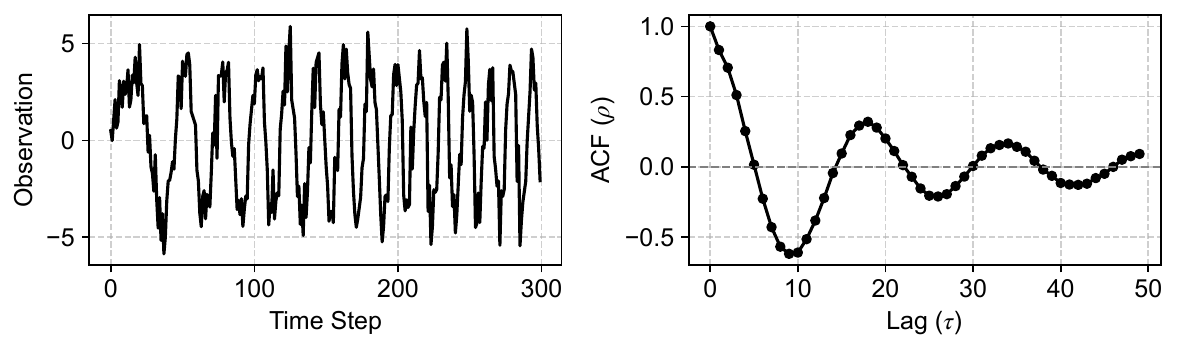}}\hfill
    \subfigure[Time-series with seasonal pattern and its ACF.]{\includegraphics[width=0.48\textwidth]{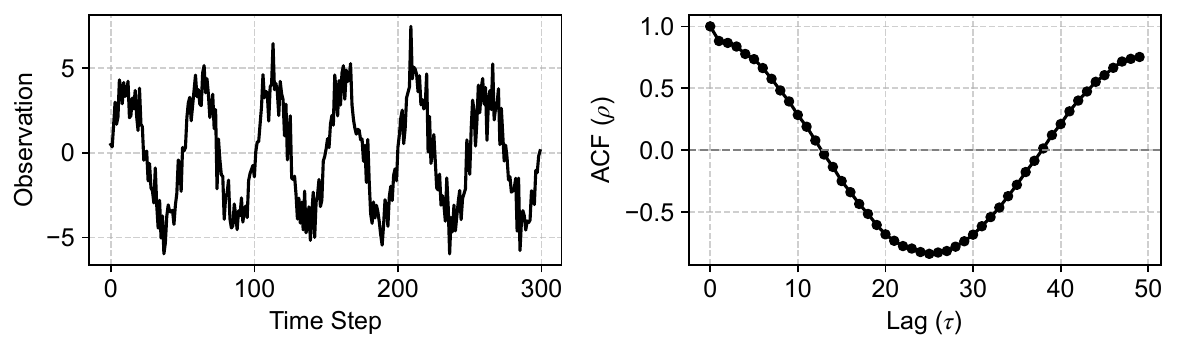}}
    \caption{The visualization of representative temporal patterns and their ACFs.}
    \label{fig:case1}
\end{figure*}

\section{Preliminaries}\label{sec:preliminaries}
\subsection{Problem Formulation}

In this paper, uppercase bold letters (\eg $\mathbf{X}$) denote matrices, lowercase bold letters (\eg $\mathbf{x}$) denote vectors, and lowercase regular letters (\eg $x$) denote scalars.
A time-series dataset is a sequence of ordered observations, denoted as $\mathbf{s} = [s_1, s_2, \ldots, s_\mathrm{M}] \in \mathbb{R}^{\mathrm{M}}$, where $\mathrm{M}$ is the number of observations.
Notably, we use a univariate formulation since covariate interaction is not the focus of this survey.
The formulation can be naturally extended to multivariate sequences~\cite{wen2025fftsurvey}.

At each time step $n$, the task is defined by four elements:
\begin{enumerate}[leftmargin=*,label=(\arabic*)]
    \item \textbf{History sequence:} $\mathbf{x} = [s_{n-\mathrm{H}+1}, \ldots, s_n] \in \mathbb{R}^{\mathrm{H}}$, where $\mathrm{H}$ denotes the history length.
    \item \textbf{Label sequence:} $\mathbf{y} = [s_{n+1}, \ldots, s_{n+\T}] \in \mathbb{R}^{\T}$, where $\T$ denotes the forecast horizon.
    \item \textbf{Backbone architecture:} $f_\theta$, which generates the predicted sequence $\hat{\mathbf{y}}=f_\theta(\mathbf{x})$, where $\theta$ denotes learnable parameters.
    \item \textbf{Loss function:} $\mathcal{E}$, which quantifies the discrepancy between the label sequence $\mathbf{y}$ and the predicted sequence $\hat{\mathbf{y}}$.
\end{enumerate}
The task of time-series forecasting is to train $f_\theta$ by minimizing $\mathcal{E}$, so that the generated $\hat{\mathbf{y}}$ accurately approximates $\mathbf{y}$~\cite{wang2025nipstimeo1,wen2023transformersurvey}.

Time-series forecasting tasks can be categorized along two dimensions~\cite{deepsurvey}.
The first dimension concerns the \textit{forecast horizon}.
Short-term forecasting predicts a relatively small number of future time steps, whereas long-term forecasting predicts a relatively large number~\cite{wang2025iclrfredf,OLinear}.
The second dimension concerns the \textit{output format}.
Deterministic forecasting produces point estimates for future values, whereas probabilistic forecasting estimates the conditional predictive distribution $p(\mathbf{y}|\mathbf{x})$.
The specific task formulation is driven by the requirements and constraints of the application domain~\cite{wuk2vae,yan2024probabilistic}.

\subsection{Autocorrelation in Time-series Forecasting}
Autocorrelation refers to the dependency between an observation and its predecessors~\cite{Arima}, distinguishing time-series from the generic tabular data. 
Its strength is quantified by the autocorrelation function (ACF)~\cite{hyndman2018forecasting}, which measures the linear correlation between a time-series and its $\tau$-lagged counterpart; large absolute ACF values at non-zero lags indicate significant autocorrelation.

\begin{definition}[ACF]
For a univariate time-series denoted as $\mathbf{s} = [s_1, s_2, \ldots, s_\mathrm{M}]$, the ACF at lag $\tau$ is defined as:
\begin{equation}
    \rho(\tau) = \frac{\sum_{n=1}^{\mathrm{M}-\tau}(s_n-\bar{s})(s_{n+\tau}-\bar{s})/(\mathrm{M}-\tau)}{\sum_{n=1}^{\mathrm{M}}(s_n-\bar{s})^2 / \mathrm{M}},
\end{equation}
where $\tau$ denotes the lag, $\bar{s}$ is the sample mean of the series.
\end{definition}

Autocorrelation is a foundational property of time-series.
On the one hand, it is widely observed in real-world time-series datasets (see \autoref{fig:case2}).
On the other hand, it illustrates key patterns in time-series, including trend (\eg increasing internet traffic), seasonality (\eg daily temperature fluctuations), and cyclicity (\eg economic cycles).
A case study in \autoref{fig:case1} substantiates this claim by visualizing the ACF of these typical patterns. 
Key observations are summarized as follows.
\begin{itemize}[leftmargin=*]
    \item \textbf{White noise} consists of random observations without any predictable pattern.
Its ACF approximates zero at all nonzero lags, indicating absence of autocorrelation, \ie each observation is decorrelated with its predecessors.
    \item \textbf{Trend} manifests as a persistent increase or decrease over time. It is often reflected by consistently positive or negative ACF across successive lags.
    \item \textbf{Cyclicity} exhibits recurrent but irregular rises and falls over time. It is reflected by alternating ACF values that do not necessarily follow a fixed period.
    \item \textbf{Seasonality} is a special case of cyclicity where fluctuations recur with a fixed period~\cite{hyndman2018forecasting}.
    Therefore, it is also termed periodicity.
    It is indicated by pronounced peaks of ACF at lags that are integer multiples of the seasonal period.
\end{itemize}

Given its foundational role, autocorrelation is critical to effective time-series forecasting.
Classical statistical approaches have long recognized its importance~\cite{anderson2011statistical,hyndman2018forecasting}, establishing formal methods for its identification (\eg  Durbin-Watson test~\cite{durbin1950testing}), correction (\eg  Cochrane-Orcutt transform~\cite{cochrane1949application}), and exploitation (\eg  ARIMA~\cite{Arima}, VAR~\cite{VAR}, and GLS~\cite{fgls}).
In deep time-series forecasting, autocorrelation remains critical, as it is inherent to both the input and label sequences.
Consequently, two research questions arise~\cite{wang2025nipstimeo1,wang2025iclrfredf}: \textbf{\ding{182} \textit{how to devise backbone architectures to accommodate autocorrelation in the history sequence}}, and \textbf{\ding{183} \textit{how to devise loss functions to accommodate autocorrelation in the label sequence}.}

Real-world time-series often arise from nonlinear dynamical systems~\cite{kantz2003nonlinear}, which has two important implications.
First, ACF, which characterizes linear temporal dependencies, cannot fully capture temporal dependence: even a series with zero ACF at every nonzero lag may exhibit nonlinear temporal dependencies.
Second, nonlinear models can capture temporal dependencies beyond the representational limits of linear models.
Classical examples include radial basis functions~\cite{sheta2001time}, local linear fits \cite{dudek2016pattern}, and polynomials~\cite{su2010prediction}, which do not restrict predictions to a fixed linear combination of past observations.
Neural networks offer a flexible, data-driven approach to learning such nonlinear temporal dependence, making them a natural choice when nonlinear temporal dependence matters for forecasting.

\section{Taxonomy}\label{sec:taxonomy}

\definecolor{fill_0}{RGB}{246, 249, 255}
\definecolor{fill_1}{RGB}{255, 246, 238}
\definecolor{fill_2}{RGB}{244, 249, 241}

\definecolor{draw_0}{RGB}{54, 110, 210}
\definecolor{draw_1}{RGB}{237, 125, 49}
\definecolor{draw_2}{RGB}{112, 173, 71}

\definecolor{fill_leaf}{RGB}{248, 248, 248}
\definecolor{draw-leaf}{RGB}{135, 135, 135}

\tikzstyle{my-box}=[
    rectangle,  
    draw=draw-leaf,
    rounded corners,
    text opacity=1,
    minimum height=1.5em,
    minimum width=5em,
    inner sep=2pt,
    align=center,
    fill opacity=.5,
    line width=0.8pt,
]
\tikzset{
leaf/.style={
my-box,
minimum height=1.5em,
fill=fill_leaf, 
text=black,
align=left,
font=\small,
inner xsep=2pt,
inner ysep=4pt,
line width=1pt
}
}
\begin{figure*}[!t]
\centering
\begin{adjustbox}{width=1\textwidth}
\begin{forest}
forked edges,
for tree={
    grow=east,
    reversed=true,
    anchor=base west,
    parent anchor=east,
    child anchor=west,
    base=center,
    font=\large,
    rectangle,
    draw=draw-leaf,
    rounded corners,
    align=left,
    text centered,
    minimum width=5em,
    edge+={darkgray, line width=1pt},
    s sep=10pt,
    l sep=10pt,
    inner xsep=2pt,
    inner ysep=3pt,
    line width=1.2pt,
    ver/.style={rotate=90, child anchor=north, parent anchor=south, anchor=center},
},
where level=0{text width=18em, fill=fill_0, draw=draw_0, font=\large, align=center, rotate=90, parent anchor=south, xshift=-25mm}{},
where level=1{text width=12em,fill=fill_1,draw=draw_1,font=\large,align=center}{},
where level=2{text width=12em,fill=fill_2,draw=draw_2,font=\large,align=center}{},
where level=3{font=\normalsize,}{},
[
    Deep Time-series Forecasting,
    [ 
        Backbone \\Architectures (Sec~\ref{sec:architecture})
        [
            Non-Transformers \\ (Sec~\ref{sec:nontrans})
            [
                \textbf{\textit{RNN}} (Sec~\ref{sec:rnn}): P-sLSTM~\cite{P-sLSTM}{,} DSTMamba~\cite{dstmamba}{,} MixMamba~\cite{MixMamba}{,} TVFSSM~\cite{TVFSSM}{,} SpaceTime~\cite{SpaceTime}{,} Mamba~\cite{gu2023mamba}{,} LRU~\cite{LRU}{,} \\
                 S4~\cite{S4}{,}  LMU~\cite{LMU}{,} DeepAR~\cite{DeepAR}{,} DeepSSM~\cite{DeepSSM}{,} SRU~\cite{SRU}{,} GRU~\cite{GRU}{,} LSTM~\cite{LSTM},  leaf, text width=50em
            ]
            [
                \textbf{\textit{CNN}} (Sec~\ref{sec:cnn}): WaveTS~\cite{WaveTS}{,} AdaWaveNet~\cite{yu2024adawavenet}{,} WFTNet~\cite{WFTNet}{,} ModernTCN~\cite{Moderntcn}{,} MICN~\cite{MICN}{,} TimesNet~\cite{Timesnet}{,}  SCINet~\cite{SCINet}{,} \\
                 DESCINet~\cite{Descinet}{,} FiLM~\cite{Film}{,} FilterNet~\cite{Filternet}{,} TCN~\cite{tcn}{,} WaveNet~\cite{wavenet}, leaf, text width=50em
            ]
            [
                \textbf{\textit{DNN}} (Sec~\ref{sec:dnn}): TimeMixer++~\cite{wang2025timemixer++}{,} SparseTSF~\cite{SparseTSF}{,} WPMixer~\cite{WPmixer}{,} CycleNet~\cite{lin2024cyclenet}{,} RLinear~\cite{RLinear}{,} TimeMixer~\cite{wang2024timemixer}{,}  FITS~\cite{FITS}{,} \\
                 MoLE~\cite{MoLE}{,} RPMixer~\cite{RPMixer}{,} FreTS~\cite{FreTS}{,} TSMixer~\cite{TSMixer}{,} TiDE~\cite{TiDE}{,} DLinear~\cite{DLinear}, leaf, text width=50em
            ]
        ]
        [
            Transformers \\(Sec~\ref{sec:trans})
            [
                \textbf{\textit{Standard self-attention models}} (Sec~\ref{sec:standardselfattention}): Timer-XL~\cite{Timer-XL}{,} Sundial~\cite{Sundial}{,} LLM4TS~\cite{LLM4TS}{,} TEMPO~\cite{Tempo}{,} GPT4MTS~\cite{GPT4mts}{,} 
                Moirai~\cite{Moirai}{,} \\  UniTS~\cite{UniTS}{,} Timer~\cite{Timer}{,} TimesFM~\cite{TimesFM}{,} Chronos~\cite{Chronos}{,} S$^2$IP-LLM~\cite{S2IP-LLM}{,} Time-FFM~\cite{liu2024time-ffm}{,} Time-LLM~\cite{Time-LLM}{,}
                 PatchTST~\cite{PatchTST}{,} \\ LLMTime~\cite{LLMTime}{,} PromptCast~\cite{PromptCast}{,}  GPT4TS~\cite{GPT4TS}, leaf, text width=50em
            ]
            [
                \textbf{\textit{Modified attention models}} (Sec~\ref{sec:modifiedattention}): TQNet~\cite{tqnet}{,} TimeBridge~\cite{liu2024timebridge}{,} Freeformer~\cite{Freeformer}{,} Fredformer~\cite{fredformer}{,} PDF~\cite{pdf}{,} \\
                DeformableTST~\cite{DeformableTST}{,} SAMformer~\cite{SAMformer}{,} iTransformer~\cite{itransformer}{,} AttentionMixer~\cite{wang2024taiattentionmixer}{,} Crossformer~\cite{Crossformer}{,} Fedformer~\cite{fedformer}{,} \\
                Pyraformer~\cite{Pyraformer}{,} Informer~\cite{Informer}{,} Autoformer~\cite{Autoformer}{,}  LogTrans~\cite{LogTrans}{,} DSANet~\cite{DSANet}, leaf, text width=50em
            ]
        ]
        [
            Plug-in Layers \\ (Sec~\ref{sec:plugin})
            [
                \textbf{\textit{Normalization layers}} (Sec~\ref{sec:normalization}): SIN~\cite{sin}{,} FAN~\cite{fan}{,} DDN~\cite{ddn}{,}  Dish-TS~\cite{dish-ts}{,} SAN~\cite{san}{,} RevIN~\cite{RevIN}, leaf, text width=50em
            ]
            [
                \textbf{\textit{Decomposition layers}} (Sec~\ref{sec:decomposition}): Seasonal-trend decomposition~\cite{Autoformer,xPatch,patchmlp,Tempo,MICN}{,} periodic decomposition~\cite{Times2d,Timesnet}{,} \\ resolution decomposition~\cite{SCINet,Descinet,wang2024timemixer,wang2025timemixer++}, leaf, text width=50em
            ]
            [
                \textbf{\textit{Tokenization layers}} (Sec~\ref{sec:tokenization}): SRSNet~\cite{wusrsnet}{,} WaveToken~\cite{WaveToken}{,} PatchMLP~\cite{patchmlp}{,} SimpleTM~\cite{Simpletm}{,} ElasTST~\cite{ElasTST}{,} MTST~\cite{MTST}{,} \\ TimeSQL~\cite{timesql}{,} PatchTST~\cite{PatchTST} 
                , leaf, text width=50em
            ]
        ]
    ]
    [ 
        Loss\\Functions (Sec~\ref{sec:loss}) 
        [
            Likelihood\\Estimation (Sec~\ref{sec:likelihoodestimation})
            [
                \textbf{\textit{Label transformation}} (Sec~\ref{sec:labeltransform}):  LatentTSF~\cite{yang2026observations}{,} Time-o1~\cite{wang2025nipstimeo1}{,}  FreDF~\cite{wang2025iclrfredf}{,} DBLoss~\cite{qiudbloss}{,} TDAlign~\cite{tdalign}{,} 
                TimeSQL~\cite{timesql}{,} \\AutoMSE~\cite{AutoMSE}, leaf, text width=50em
            ]
            [
                \textbf{\textit{Covariance modeling}} (Sec~\ref{sec:covariancemodeling}): MMKE~\cite{MMKE}{,} MKE~\cite{MKE}{,} QDF~\cite{wang2026iclrqdf}
                , leaf, text width=50em
            ]
        ]
        [
            Shape \\ Alignment (Sec~\ref{sec:shapealignment})
            [
            \textbf{\textit{Dynamic time warping}} (Sec~\ref{sec:shapealignment}): STRIPE~\cite{STRIPE2}{,} GDTW~\cite{GDTW}{,} DSDTW~\cite{DSDTW}{,} GromovDTW~\cite{GDTW2}{,} Dilate~\cite{Dilate}{,} ShapeDTW~\cite{ShapeDTW}{,} \\ LDTW~\cite{LDTW}{,}  SoftDTW~\cite{soft-dtw}{,} 
            DTW~\cite{dtw} 
            , leaf, text width=50em
            ]
        ]
        [
            Distribution\\Balancing (Sec~\ref{sec:distributionbalancing})
            [
                \textbf{\textit{Discrepancy minimization}} (Sec~\ref{sec:discrepancyminimization}): 
                KMB-DF~\cite{pan2026kmbdf}{,} TimeAlign~\cite{hu2026bridging}{,}
                DistDF~\cite{wang2026iclrdistdf}{,} PSLoss~\cite{psloss}{,} PSW~\cite{wang2025iclrpswi}
                , leaf, text width=50em
            ]
            [
                \textbf{\textit{Adversarial training}} (Sec~\ref{sec:adversarialbalancing}):  AttnWGAIN~\cite{AttnWGAIN}{,} SDGCN~\cite{SDGCN}{,} WRCGAN~\cite{WRCGAN}{,} TrendGCN~\cite{TrendGCN}{,} AST~\cite{AST}
                , leaf, text width=50em
            ]
        ]
        [
            Conditional\\Generation (Sec~\ref{sec:conditionalgeneration})
            [
                \textbf{\textit{Diffusive generation}} (Sec~\ref{sec:diffusion}):   StochDif~\cite{StochDif}{,} SSD-TS~\cite{SSD-TS}{,} Diffusion-TS~\cite{Diffusion-TS}{,} D3U~\cite{D3U}{,} TimeDart~\cite{TimeDart}{,}
                NSDiff~\cite{NSDiff}{,} \\ CNDiff~\cite{CNDiff}{,} TCDM~\cite{TCDM}{,} 
                TimeWeaver~\cite{TimeWeaver}{,} D3M~\cite{D3M}{,} TMDM~\cite{TMDM}{,} MG-TSD~\cite{MG-TSD}{,} 
                TimeDiff~\cite{TimeDiff}{,} SSSD~\cite{SSSD}{,}
                D3VAE~\cite{D3VAE}{,} \\
                CSDI~\cite{CSDI}{,}  TimeGrad~\cite{TimeGrad}   
                , leaf, text width=50em
            ]
            [
                \textbf{\textit{Autoregressive generation}} (Sec~\ref{sec:autoregressivegeneration}):  Timer-XL~\cite{Timer-XL}{,} TimeBase~\cite{Timebase}{,} MoLA~\cite{MoLA}{,} LangTime~\cite{LangTime}{,} 
                AutoTimes~\cite{Autotimes}{,} Timer~\cite{Timer}{,} \\ DeepAR~\cite{DeepAR}{,} LSTNet~\cite{LSTNet}
                , leaf, text width=50em
            ]
            ]
    ]
]
\end{forest}
\end{adjustbox}
\vspace{-2mm}
\caption{The proposed taxonomy surrounding autocorrelation modeling. The literature in each branch is sorted by recency.}
\label{fig:overview_tree}
\end{figure*}

This section introduces existing taxonomies and compares them with the proposed taxonomy.
By way of preface, one key distinction warrants emphasis: \textit{this survey focuses on deep time-series forecasting literature that models autocorrelation to improve forecasting performance.}
While topics such as denoised learning~\cite{zhou2024denoising} and channel dependency~\cite{zhao2024channel,han2024channel,qiuduet} are valuable, they are not unique challenges in time-series forecasting and fall outside the scope of this survey.

\subsection{Taxonomy in Related Surveys}\label{sec:taxonomyrelatedsurveys}
Previous surveys mainly organize the literature along three dimensions.
The first and most common dimension is backbone architecture~\cite{tsinghuasurvey,wen2024foundationsurvey,deepsurvey}.
For example, Li et al.~\cite{li2024deep} categorize models into CNNs, RNNs, graph neural networks (GNNs), and Transformers.
Wang et al.~\cite{tsinghuasurvey} additionally consider DNNs and plug-in layers, whereas Casolaro et al.~\cite{casolaro2023deep} further consider deep Gaussian processes and generative models. 
Along this line, several surveys focus exclusively on specific model families, such as RNNs~\cite{fang2021rnnsurvey} and Transformers~\cite{wen2023transformersurvey}.
The second dimension is task formulation.
For example, Benidis et al.~\cite{acmsurvey} distinguish between deterministic and probabilistic forecasting.
The third dimension concerns the application of general learning techniques to forecasting, including self-supervised learning~\cite{selfsurvey}, data augmentation~\cite{wen2021augmentsurvey}, and diffusion models~\cite{wen2024diffsurvey,lin2024diffusionsurvey}.
These surveys organize literature according to the taxonomy of the underlying technique.
For example, Zhang et al.~\cite{selfsurvey} categorize literature based on their contrastive learning strategies.
Notably, Jadon et al.~\cite{jadon2024losssurvey} discuss loss functions, but they only cover general-purpose losses (\eg hinge loss), ignoring loss functions specifically designed for time-series forecasting.
A summary is presented in Table~\ref{tab:gen_works}.

\subsection{Taxonomy in This Survey}\label{sec:taxonomythissurvey}

The proposed taxonomy is presented in \autoref{fig:overview_tree}.
At the top level, it divides the literature into two branches: \textit{backbone architectures} and \textit{loss functions}, which address challenges \ding{182} and \ding{183} posed by autocorrelation, respectively.
The next level groups methods according to how they model autocorrelation.
The backbone architecture branch includes non-Transformer and Transformer-based models, as well as plug-in layers compatible with different architectures.
The loss function branch comprises four categories, each representing a distinct strategy for designing loss functions that account for label autocorrelation.

The proposed taxonomy differs from existing taxonomies in two respects. 
First, it centers on \textit{autocorrelation modeling}, providing a unified and focused framework for understanding how methods have evolved to model this fundamental property of time-series.
Second, it systematically discusses \textit{loss functions}, which receive little attention in previous surveys.

\begin{table*}
\centering
\caption{Overview of recent surveys. }
\label{tab:gen_works}
\scriptsize
\renewcommand{\arraystretch}{1} 
\begin{tabular}{
    >{\raggedright\arraybackslash}p{2.2cm} 
    >{\raggedright\arraybackslash}p{1cm}   
    >{\raggedright\arraybackslash}p{2.5cm} 
    >{\centering\arraybackslash}p{2.1cm}   
    >{\centering\arraybackslash}p{1.3cm} 
    >{\centering\arraybackslash}p{1.2cm}    
    >{\centering\arraybackslash}p{0.9cm} 
    >{\centering\arraybackslash}p{1.0cm} 
    >{\centering\arraybackslash}p{1.0cm} 
    >{\centering\arraybackslash}p{1.0cm} %
}
\toprule
\multirow{2}[2]{*}{\textbf{Author}} & \multirow{2}[2]{*}{\textbf{Year}} & \multirow{2}[2]{*}{\textbf{Taxonomy}} & \multicolumn{3}{c}{\textbf{Backbone Architecture}} & \multicolumn{4}{c}{\textbf{Loss Function}} \\ 
\cmidrule(lr){4-6} \cmidrule(lr){7-10}
 &&& \textbf{Non-Transformers} & \textbf{Transformers} & \textbf{Plug-ins} & \textbf{LikeEst} & \textbf{ShapeAlign} & \textbf{DistBal} & \textbf{CondGene}\\
\midrule
\rowcolor{lightgray} Wen et al.~\cite{wen2023transformersurvey} & 2023 & Backbone architecture & \false & \true & \false &  \false & \false & \false & \false\\
Li et al.~\cite{li2024deep} & 2024 & Backbone architecture & \true & \true & \false &  \false & \false & \false & \false\\
\rowcolor{lightgray} Wang et al.~\cite{tsinghuasurvey} & 2024 & Backbone architecture & \true & \true & \true &  \false & \false & \false & \false\\
Casolaro et al.~\cite{casolaro2023deep} & 2023 & Backbone architecture & \true & \true & \false &  \false & \false & \false & \true\\
\rowcolor{lightgray} Liang et al.~\cite{wen2024foundationsurvey} & 2024 & Backbone architecture & \true & \true & \false &  \false & \false & \false & \true \\
Yi et al.~\cite{wen2025fftsurvey} & 2025 & Backbone architecture & \true & \true & \false &  \false & \false & \false & \false \\
\rowcolor{lightgray} Benidis et al.~\cite{acmsurvey} & 2022 & Task formulation & \true & \true & \false &  \false & \false & \false & \false \\
Zhang et al.~\cite{selfsurvey} & 2024 & Self-supervised learning & \false & \false & \false &  \false & \false & \false & \true \\
\rowcolor{lightgray} Yang et al.~\cite{wen2024diffsurvey} & 2024 & Diffusion models & \false & \false & \false &  \false & \false & \false & \true \\
Lin et al.~\cite{lin2024diffusionsurvey} & 2024 & Diffusion models & \false & \false & \false &  \false & \false & \false & \true \\
\rowcolor{lightgray} Jadon et al.~\cite{jadon2024losssurvey} & 2024 & General loss functions & \false & \false & \false &  \false & \false & \false & \false\\
This survey & 2026 & Autocorrelation & \true & \true & \true & \true & \true & \true & \true \\
\bottomrule
\end{tabular}
\begin{tablenotes}
    \item \textit{Note:} ``LikeEst'' abbreviates likelihood estimation,  ``ShapeAlign'' abbreviates shape alignment, ``DistBal'' abbreviates distribution balancing, ``CondGene'' refers to conditional generation.
\end{tablenotes}
\end{table*}

\section{Backbone Architectures for Modeling Autocorrelation}\label{sec:architecture}
\begin{figure*}
    \centering
    \subfigure[RNN-based models.]{\includegraphics[width=0.31\textwidth]{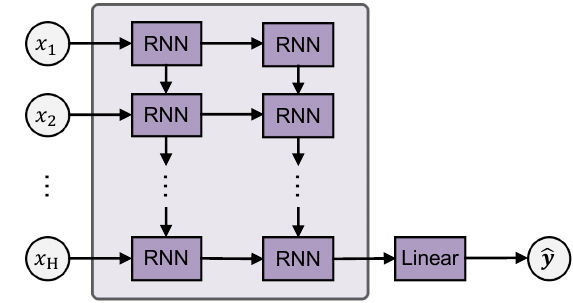}}\hfill
    \subfigure[CNN-based models.]{\includegraphics[width=0.31\textwidth]{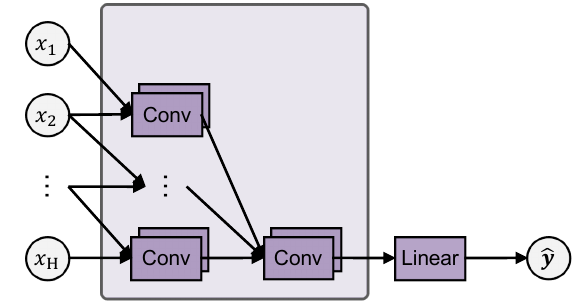}}\hfill
    \subfigure[DNN-based models.]{\includegraphics[width=0.31\textwidth]{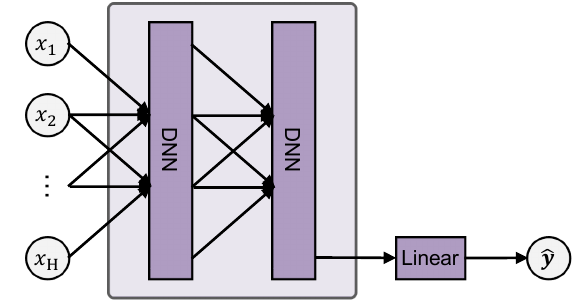}}\hfill
    \caption{Overview of representative non-Transformer architectures.
Circular nodes represent model inputs and outputs.}
    \label{fig:diagram1}
\end{figure*}

In deep time-series forecasting, the key challenge of devising backbone architectures is modeling \textit{history autocorrelation}, \ie the autocorrelation within the history sequence.
Suboptimal modeling of history autocorrelation produces impoverished representation of $\mathbf{x}$ and degrades forecasting performance.

Most backbone architectures adopt the direct forecasting (DF) paradigm~\cite{itransformer,Timesnet,lin2024cyclenet}, generating the full predicted sequence $\hat{\mathbf{y}}$ in a single forward pass:
\begin{equation}\label{eq:df}
\hat{\mathbf{y}}=f_\theta(\mathbf{x})=f_\mathrm{linear}\circ f_\mathrm{nn}(\mathbf{x}),
\end{equation}
where $f_\mathrm{nn}$ extracts representations from $\mathbf{x}$, which are then projected by a linear layer to produce $\hat{\mathbf{y}}\in\mathbb{R}^{\T}$.
Backbones differ primarily in the design of $f_\mathrm{nn}$ and fall into two broad categories: non-Transformer models (Section~\ref{sec:nontrans}) and Transformer-based models (Section~\ref{sec:trans}).
Complementing these architectures, plug-in layers are developed to improve history autocorrelation modeling across different backbone architectures (Section~\ref{sec:plugin}).

\subsection{Non-Transformer Models}\label{sec:nontrans}
This subsection reviews non-Transformer models, including RNNs, CNNs, and DNNs.
Recent advancements such as state space models~\cite{gu2023mamba}, large-kernel CNNs~\cite{Moderntcn}, and deep decomposition models~\cite{wang2024timemixer} are covered, which  better models complex history autocorrelation patterns.

\subsubsection{RNNs}\label{sec:rnn}
These architectures are natural choice to model sequential data such as video~\cite{rnnvideo}, texts~\cite{peng2023rwkv} and logs~\cite{pan2025cmoe}.
In time-series, RNNs model autocorrelation as the dependency between successive hidden states~\cite{grud}.
Given an input sequence $\mathbf{x} = [x_1, x_2, \ldots, x_\mathrm{H}] \in \mathbb{R}^{\mathrm{H}}$, the hidden state $\mathbf{z}_t \in \mathbb{R}^\mathrm{D}$, with hidden dimension $\mathrm{D}$, evolves as:
\begin{equation*}\label{eq:state_space_model}
    \begin{aligned}
    \mathbf{z}_t &= \mathbf{A} \mathbf{z}_{t-1} + \mathbf{b} x_t,\\
    \hat{x}_{t+1} &= \mathbf{c}^\top \mathbf{z}_t + d,
    \end{aligned}
\end{equation*}
where $\mathbf{A}$, $\mathbf{b},\mathbf{c}$, and $d$ are learnable parameters, and $\hat{x}_{t+1}$ is the next-step prediction.
For $t > \mathrm{H}$, the backbone iteratively rolls out by setting $x_t\leftarrow\hat{x}_t$.
This formulation aligns with the \textit{state space model (SSM)}, where  $\mathbf{A}$ governs how past information propagates through the hidden state.
Introducing a nonlinear activation $\sigma(\cdot)$ gives a vanilla RNN:
\begin{equation*}\label{eq:vanilla_rnn_model}
    \begin{aligned}
        \mathbf{z}_t &= \sigma(\mathbf{A} \mathbf{z}_{t-1} + \mathbf{b} x_t),\\
        \hat{x}_{t+1} &= \mathbf{c}^\top \mathbf{z}_t + d,
        \end{aligned}
\end{equation*}
where the model parameters are shared across time steps, allowing RNNs to process variable-length sequences.
However, RNNs face two limitations in time-series forecasting.
\begin{itemize}[leftmargin=*]
    \item \textbf{Limited Efficiency}.
The sequential update of $\mathbf{z}_t$ precludes parallelization, leading to high latency given long sequences.
Subsequent architectures incorporate linearized or simplified recurrences (\eg LRU~\cite{LRU} and SRU~\cite{SRU}), which enables parallel computation and improves efficiency.
    \item \textbf{Memory Bottlenecks}.
The hidden state $\mathbf{z}_t$ is expected to memorize all history information relevant for prediction, as it is the sole basis for forecasting.
However, retaining all relevant information is difficult because hidden states have finite capacity, which bottleneck long-range autocorrelation modeling.
Early solutions such as LSTM~\cite{LSTM} and GRU~\cite{GRU} introduce memory states and employ gating mechanisms to selectively update them.
Notable examples include DeepAR~\cite{DeepAR}, which employs LSTM to encode the history sequence, and P-sLSTM, which modernizes LSTM with an exponential gating mechanism~\cite{P-sLSTM}.
However, these heuristic designs do not guarantee retention of all history information relevant to prediction.
An alternative line of work constructs compact memory states by projecting history onto polynomial bases.
This direction, pioneered by LMU~\cite{LMU} and formalized by HiPPO~\cite{hippo}, provides a principled approximation to the history and underpines subsequent improvements~\cite{hippo2,hippo3}.
\end{itemize}

Deep SSMs have recently emerged as a promising approach to addressing both limitations~\cite{S4,DeepSSM,TVFSSM,SpaceTime}.
An exemplar is Mamba~\cite{gu2023mamba}, which enhances memory through a parallel-scan algorithm. 
Its effectiveness has motivated adaptations for time-series forecasting~\cite{smamba,affirmmamba}.
For example, DSTMamba~\cite{dstmamba} employs Mamba to model seasonal patterns in traffic prediction; MixMamba~\cite{MixMamba} integrates Mamba with mixture-of-experts to improve non-stationary time-series forecasting.

\subsubsection{CNNs}\label{sec:cnn}
These architectures are originally designed for images to model correlations among pixels.
They are adapted for time-series forecasting by treating sequences as one-dimensional images~\cite{tcn}.
In this context, autocorrelation is modeled as the correlation between adjacent time steps within a local receptive field.
Given a history sequence $\mathbf{x} \in \mathbb{R}^{\mathrm{H}}$, the hidden state $\mathbf{z}_t \in \mathbb{R}^\mathrm{D}$, with $\mathrm{D}$ output features, is computed as
\begin{equation*}
    \begin{aligned}
        \mathbf{z}_t &= \mathbf{w}_1 x_{t-\kappa+1} + \mathbf{w}_2 x_{t-\kappa+2} + \cdots + \mathbf{w}_{\kappa} x_t,
    \end{aligned}
\end{equation*}
where $\kappa$ is the kernel size and $\mathbf{w}_1, \ldots, \mathbf{w}_{\kappa}\in\mathbb{R}^\mathrm{D}$ are convolution kernel weights shared across all time steps.
The popularity of CNNs in this field is catalyzed by TCN~\cite{tcn}, compromising stacked causal convolutions with residual connections, which outperforms traditional RNNs (\eg LSTM~\cite{LSTM}, GRU~\cite{GRU}). 

A key challenge for CNNs is modeling long-term autocorrelation.
This challenge can be interpreted as expanding the receptive field~\cite{ding2024unireplknet}, \ie the range of input steps influencing the output.
Current models expanding receptive field can be categorized by their operation domains as follows.
\begin{itemize}[leftmargin=*]
    \item \textbf{Time-domain models.}
They expand receptive field by modifying the convolution operator.
One line of work employs \textit{dilated convolutions} (\eg TCN~\cite{tcn} and WaveNet~\cite{van2016wavenet}), which expand receptive field exponentially without increasing parameters.
A second line of work employs \textit{iterative downsampling} strategies (\eg SCINet~\cite{SCINet}), which partition $\mathbf{x}$ into sub-sequences and process them with separate convolution layers.
To mitigate information loss caused by downsampling, DESCINet~\cite{Descinet} introduces residual connections to enable information exchange between sub-sequences; TimesNet~\cite{Timesnet} and MICN~\cite{MICN} perform multiscale fusion to aggregate information from intermediate sampling layers.
A third line of work employs \textit{large convolutional kernels}. Inspired by their success in vision tasks~\cite{ding2024unireplknet}, ModernTCN~\cite{Moderntcn} adapts large-kernel depth-wise convolutions to time-series forecasting.
    \item \textbf{Frequency-domain models.}
According to the convolution theorem~\cite{oppenheim1997signals}, a frequency-domain elementwise multiplication (\ie \textit{frequency-domain filtering}) equals a time-domain convolution with kernel size equal to the sequence length, making frequency-domain operations a promising means of expanding receptive field.
For instance, FiLM~\cite{Film} transforms the input to the frequency domain, applies a learnable frequency-domain filter, and then transforms it back to the time domain.
FilterNet~\cite{Filternet} further enables adaptive filtering through input-dependent filters.
Subsequent studies incorporate wavelet transform to capture time-varying frequency characteristics, improving robustness against non-stationarity~\cite{yu2024adawavenet,WaveTS,WFTNet}.
\end{itemize}

\subsubsection{DNNs}\label{sec:dnn}
These architectures are originally developed for modeling general tabular data with minimal inductive bias, adapted to this task by treating the flattened $\mathbf{x}$ as a feature vector.
This direction is notably popularized by DLinear~\cite{DLinear}, which demonstrates that even a simple dense linear layer can effectively model autocorrelation and exhibit competitive performance.
Specifically, a basic dense linear layer maps the history $\mathbf{x}\in\mathbb{R}^{\mathrm{H}}$ to the predictions as:
\begin{equation*}
    \hat{\mathbf{y}} = \mathbf{W}\mathbf{x},
\end{equation*}
where $\hat{y}_t\in\hat{\mathbf{y}}$ aggregates all historical observations with learnable weights, adaptively modeling history autocorrelation with a global receptive field.
However, a single linear layer struggles to model complex autocorrelation patterns.
Current approaches address this via two strategies.
\begin{itemize}[leftmargin=*]
    \item \textbf{Simplifying autocorrelation.}
This strategy transforms data to suit simple linear models.
For instance, DLinear~\cite{DLinear} decomposes $\mathbf{x}$ into trend and seasonal components and models them by separate linear layers.
RLinear~\cite{RLinear} employs the RevIN technique~\cite{RevIN} to avoid the complexity of non-stationary autocorrelations.
CycleNet~\cite{lin2024cyclenet} learns to remove periodic patterns and then applies a linear layer to the residual.
SparseTSF~\cite{SparseTSF} decomposes $\mathbf{x}$ by periodicity and applies linear layers to each component, avoiding the complexity of coupled periods. Additionally, constructing backbone architectures in the frequency domain is a ubiquitous strategy that reduces autocorrelation, thereby enabling competitive performance of linear models (\eg  FITS~\cite{FITS}, FreTS~\cite{FreTS}, and RPMixer~\cite{RPMixer}).
    \item \textbf{Enhancing backbone capacity.}
This strategy employs more expressive backbone architecture to better model complex autocorrelations.
One approach involves replacing linear layers with MLPs~\cite{TiDE,TSMixer}.
For example, TiDE~\cite{TiDE} incorporates a channel-independent MLP block with two hidden layers and residual connections; TSMixer~\cite{TSMixer} employs MLPs for both time and channel mixing.
Subsequent works adopt multi-branch architectures~\cite{wang2024timemixer,wang2025timemixer++,WPmixer}.
For instance, TimeMixer~\cite{wang2024timemixer} downsamples $\mathbf{x}$ into multiscale subsequences and processes each with a dedicated MLP; MoLE~\cite{MoLE} introduces a mixture-of-experts mechanism, using an MLP-based router to adaptively weight the outputs of multiple linear experts.
\end{itemize}

\subsection{Transformer-based Models}\label{sec:trans}
\begin{figure}
    \includegraphics[width=\linewidth]{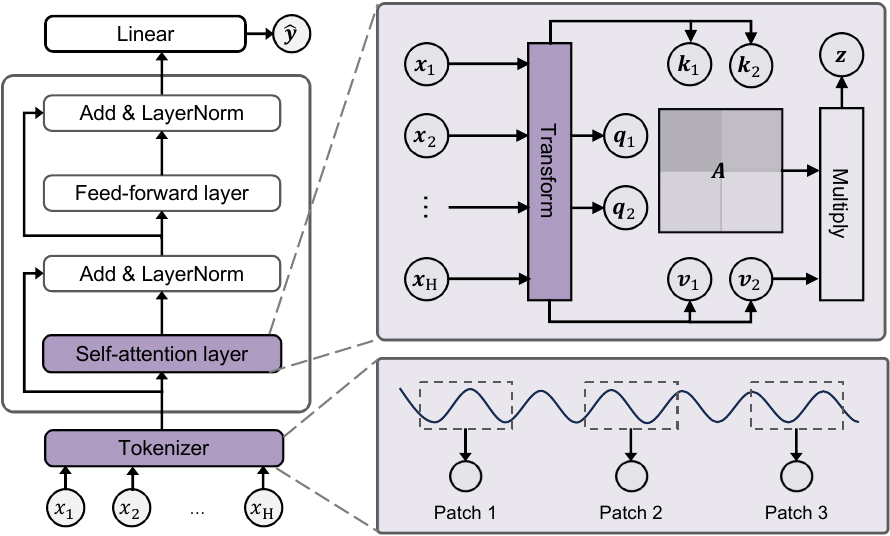}
    \caption{Overview of representative Transformer architectures.
The circlular nodes represent model input and output.
The colored blocks indicate the primary modifications to adapt Transformer to time-series forecasting.}
    \label{fig:diagram2}
\end{figure}

\begin{table*}
    \centering
    \caption{Overview of LLM-based time-series backbone architectures with standard self-attention token-mixers.}
    \label{tab:llm_models}
    \scriptsize
    \renewcommand{\arraystretch}{1} 
    \begin{tabular}{
        >{\raggedright\arraybackslash}p{1.8cm} 
        >{\raggedright\arraybackslash}p{1cm}   
        >{\raggedright\arraybackslash}p{1.5cm} 
        >{\centering\arraybackslash}p{1.1cm}   
        >{\centering\arraybackslash}p{1.1cm} 
        >{\centering\arraybackslash}p{1.1cm}    
        >{\centering\arraybackslash}p{1.1cm} 
        >{\centering\arraybackslash}p{1.1cm} 
        >{\centering\arraybackslash}p{1.1cm} 
        >{\centering\arraybackslash}p{3.0cm} 
    }
    \toprule
    \multirow{2}[2]{*}{\textbf{Model}} & \multirow{2}[2]{*}{\textbf{Year}} & \multirow{2}[2]{*}{\textbf{Backbone}} & \multirow{2}[2]{*}{\textbf{Tokenizer}} & \multirow{2}[2]{*}{\textbf{Texts}} & \multicolumn{4}{c}{\textbf{Cross-modal Alignment Strategy}} & \multirow{2}[2]{*}{\textbf{Fine-tuned LLM parameters}}\\ 
     \cmidrule(lr){6-9}
     &&&&& \textbf{Query} & \textbf{Repro} & \textbf{Pretrain}  & \textbf{Finetuning}\\
    \midrule
    \rowcolor{lightgray} PromptCast~\cite{PromptCast} & 2023 & BERT & Discrete & \true & \true & \false &  \false & \false & \false \\
    LLMTime~\cite{LLMTime} & 2023 & GPT3 & Discrete & \false & \true & \false &  \false & \false & \false \\
    \rowcolor{lightgray} Time-LLM~\cite{Time-LLM} & 2024 & LLaMA & Patching & \true  & \false &  \true & \false & \false & \false \\
    Time-FFM~\cite{liu2024time-ffm} & 2024 & GPT2 & Patching & \true  & \false &  \true & \false & \false & \false \\
    \rowcolor{lightgray} GPT4TS~\cite{GPT4TS} & 2023 & GPT2 & Patching & \false  & \false &  \false & \false & \true & LN, PE \\
    TEMPO~\cite{Tempo} & 2024 & GPT2 & Patching & \true & \false &  \false & \false & \true & LN, SA\\ 
    \rowcolor{lightgray} GPT4MTS~\cite{GPT4mts} & 2024 & GPT2 & Patching & \true & \false &  \false & \false & \true & LN, PE \\
    S$^2$IP-LLM~\cite{S2IP-LLM} & 2024 & GPT2 & Patching & \false & \false &  \true & \false & \true & LN, PE \\
    \rowcolor{lightgray} LLM4TS~\cite{LLM4TS} & 2025 & GPT2 & Patching & \false & \false &  \false & \false & \true & LN, SA, PE \\ 
    Moirai~\cite{Moirai} & 2024 & Transformer & Patching & \false & \false & \false & \true &  \false & \false \\
    \rowcolor{lightgray} UniTS~\cite{UniTS} & 2024 & Transformer & Patching & \false & \false & \false & \true &  \false & \false \\
    Timer~\cite{Timer} & 2024 & Transformer & Patching & \false & \false & \false & \true &  \false & \false \\
    \rowcolor{lightgray} Timer-XL~\cite{Timer-XL} & 2025 & Transformer & Patching & \false & \false & \false & \true &  \false & \false \\
    TimesFM~\cite{TimesFM} & 2024 & Transformer & Patching & \false & \false & \false & \true &  \false & \false \\
    \rowcolor{lightgray} Chronos~\cite{Chronos} & 2023 & T5 & Discrete & \false & \false & \false & \true &  \false & \false \\
    \bottomrule
    \end{tabular}
    \begin{tablenotes}
        \item \textit{Note:}  ``Repro'' abbreviates reprogramming, ``LN'' abbreviates LayerNorm, ``PE'' abbreviates position embedding, ``SA'' abbreviates self-attention layer. ``Prompt'' indicates the involution of textual prompts describing time-series.
    \end{tablenotes}
    \end{table*}

Transformers are originally developed for natural language processing~\cite{Transformer} and scale well on large datasets~\cite{guo2025deepseek,bert}.
In time-series forecasting, the self-attention mechanism of Transformers provides a promising approach for modeling autocorrelation through pairwise similarity~\cite{TimeXer,PatchTST}.
Given an input $\mathbf{x}\in\mathbb{R}^\mathrm{H}$, which is first embedded into $\mathbf{X}\in\mathbb{R}^{\mathrm{H}\times\mathrm{D}}$, a self-attention layer forms query, key, and value vectors and computes the attention matrix as:
\begin{equation*}
    \begin{aligned}
    \mathbf{Q} &= \mathbf{X}\mathbf{W}^\mathrm{Q}+\mathbf{1}(\mathbf{b}^\mathrm{Q})^\top,
    \mathbf{K} = \mathbf{X}\mathbf{W}^\mathrm{K}+\mathbf{1}(\mathbf{b}^\mathrm{K})^\top,\\
    \mathbf{V} &= \mathbf{X}\mathbf{W}^\mathrm{V}+\mathbf{1}(\mathbf{b}^\mathrm{V})^\top,
    \mathbf{A} = \mathrm{Softmax}(\mathbf{Q}\mathbf{K}^\top/\sqrt{\mathrm{D}}),
    \end{aligned}
\end{equation*}
where $\mathbf{W}^\mathrm{Q},\mathbf{W}^\mathrm{K},\mathbf{W}^\mathrm{V}$ and $\mathbf{b}^\mathrm{Q},\mathbf{b}^\mathrm{K},\mathbf{b}^\mathrm{V}$ are learnable parameters.
Softmax is applied row-wise, and each entry $a_{h,h'}$ reflects the similarity between time steps $h$ and $h'$.
The output at step $h$ aggregates all value vectors:
\begin{equation*}
    \mathbf{z}_h = \sum_{h'} a_{h,h'} \mathbf{v}_{h'}, \quad h=1,2,...,\mathrm{H}.
\end{equation*}

Self-attention models autocorrelation through pairwise similarities between time steps. However, individual observations often carry limited semantics without \textbf{temporal context} (\eg a temperature of 20°C may correspond to different trends depending on neighboring values)~\cite{wang2024taiattentionmixer,chencloser}, potentially compromising the reliability of attention scores.
To address this, recent works incorporate temporal context into attention scores, by integrating tokenizers~\cite{PatchTST} or modified self-attention mechanism~\cite{Autoformer,fedformer}, as illustrated in \autoref{fig:diagram2}.

\subsubsection{Standard Self-attention Models}\label{sec:standardselfattention}
These models incorporate temporal context through tokenization while retaining standard self-attention.
As illustrated in \autoref{fig:diagram2}, a typical tokenizer partitions a time-series into overlapping patches using a sliding window, treating each patch as a context-rich token to facilitate autocorrelation modeling~\cite{Time-LLM}.
Since PatchTST~\cite{PatchTST}, tokenization has become a common component in Transformer-based backbones~\cite{tan2024language,TimeXer}.

Building on the success of tokenizers, recent studies adapt LLMs for time-series forecasting~\cite{wen2024foundationsurvey,tan2024language}.
The key challenge is \textbf{representation alignment}: aligning time-series representations with the input space of LLMs, which is essential to transfer LLMs' autocorrelation modeling ability to time-series forecasting.
Existing methods follow four alignment paradigms.
\begin{itemize}[leftmargin=*]
    \item \textbf{Query-based paradigm}.
It achieves alignment by converting time-series into textual sequences and reformulating forecasting as a question-answer task~\cite{PromptCast,LLMTime}.
As a pioneering effort, PromptCast~\cite{PromptCast} introduces a representative prompt template.
This paradigm offers several advantages: it supports zero-shot forecasting, accommodates textual side information (\eg task descriptions and dataset statistics), and produces interpretable textual outputs.
Subsequent works refine prompt templates to inject domain knowledge~\cite{zhou2025context,khezresmaeilzadeh2025morfi}, adapt text tokenizers to process continuous numerical values~\cite{LLMTime, wang2025chattime}, and employ in-context learning to tailor forecasting to specific datasets~\cite{lu2024context}.

    \item \textbf{Reprogramming-based paradigm}.
It achieves alignment by translating time-series tokens into the LLM's word embeddings.
A simple strategy uses a linear projection layer to achieve this~\cite{hucontext}; during fine-tuning, only this layer is updated while the LLM remains frozen to avoid overfitting.
Another strategy employs a cross-attention layer which enables time-series tokens to query a frozen textual embedding table and retrieve relevant word embeddings (\eg  Time-LLM~\cite{Time-LLM}, Time-FFM~\cite{liu2024time-ffm}, CALF~\cite{liu2025calf}).
To improve computational efficiency, the embedding table can be compressed into a compact set of text prototypes using principal component analysis~\cite{liu2025calf} or a learnable linear mapping~\cite{Time-LLM}.
Contrastive learning has also been used to enhance alignment by maximizing the similarity between time-series tokens and selected text prototypes~\cite{liu2025calf,suntest,S2IP-LLM}.
   
    \item \textbf{Finetuning-based paradigm}.
It achieves alignment by finetuning LLM's parameters on time-series data.
GPT4TS~\cite{GPT4TS} shows that tuning only position embeddings and layer normalization parameters can yield competitive results.
This lightweight tuning strategy has been adopted in subsequent works such as GPT4MTS~\cite{GPT4mts} and S$^2$IP-LLM~\cite{S2IP-LLM}.
Later studies incorporate low-rank adaptation (LoRA) to also fine-tune self-attention layers  (\eg  LLM4TS~\cite{LLM4TS}, TEMPO~\cite{Tempo}, Time-LlaMA~\cite{Time-LlaMA}, MSFT~\cite{MSFT}).
By restricting tuning to a small set of LLM parameters, they facilitate alignment while mitigating overfitting on limited time-series data.
    \item \textbf{Pretraining-based paradigm}.
It achieves alignment by training foundation models from scratch using large-scale time-series corpora~\cite{Chronos}.
Existing models mainly include \textit{encoder-only} architectures trained with masked reconstruction (\eg  Moirai~\cite{Moirai}, UniTS~\cite{UniTS}, Moment~\cite{Moment}, Baguan-TS~\cite{Baguan-TS}), and \textit{decoder-only} architectures trained with next-step prediction (\eg  TimesFM~\cite{TimesFM}, Timer~\cite{Timer}, Time-MoE~\cite{Time-MoE}, Timer-XL~\cite{Timer-XL}).
For example, Baguan-TS~\cite{Baguan-TS} employs an \textit{encoder-only} architecture and trains it with synthetic and real-world time-series data, which effectively activates sequence-native in-context learning ability and yields consistent performance gains.
Chronos~\cite{Chronos} employs an \textit{encoder-decoder} architecture with quantization-based tokenization, trained with next-step prediction task. Sundial~\cite{Sundial} employs a decoder-only architecture trained with flow matching, and releases a pretraining corpus exceeding one trillion time points.
\end{itemize}

\subsubsection{Modified Attention Models}\label{sec:modifiedattention}
These models refine the self-attention mechanism to incorporate temporal context.
As shown in \autoref{fig:diagram2}, it is often achieved by applying time- or frequency-domain transformations before computing attention scores.
\begin{itemize}[leftmargin=*]
    \item \textbf{Time-domain transformations.}
    These methods incorporate temporal context through three main strategies.
(i) \textit{Convolution-based transforms} use learnable convolution layers to embed temporal context into query and key vectors~\cite{LogTrans,Informer,DSANet}.
For instance, LogTrans~\cite{LogTrans} uses causal convolutions for this purpose, demonstrating that convolution helps attention scores identify relevant subsequences.
(ii) \textit{Downsampling-based transforms} selectively downsample the attention matrix to enable selective integration~\cite{Pyraformer,Informer}.
Luo et al.~\cite{DeformableTST} observe that standard self-attention often produces nearly uniform scores due to temporal redundancy, motivating downsampling mechanisms.
Early methods employ heuristic downsampling strategies, such as retaining neighboring keys~\cite{Pyraformer} or selecting queries with non-uniform attention distributions~\cite{Informer}.
Subsequent works attempt to learn downsampling strategy from data: DeformableTST~\cite{DeformableTST} uses deformable attention to sample informative keys; TimeBridge~\cite{liu2024timebridge} uses MLP to downsample adjacent queries.
(iii) \textit{Transpose-based transforms} treat individual series as tokens, shifting attention from inter-step to inter-series correlations~\cite{Crossformer,itransformer,SAMformer,wang2024taiattentionmixer}.
In this way, linear projections of each series produce queries, keys, and values that encode series-wise temporal context.

    \item \textbf{Frequency-domain transformations.}
These methods integrate temporal context in spectral components.
For example, Autoformer~\cite{Autoformer} computes lagged cross-correlations via frequency-domain operations, using them to aggregate lagged value vectors.
FEDformer~\cite{fedformer} performs self-attention directly on frequency-domain components.
Freeformer~\cite{Freeformer} further adds an attention sparsity penalty to exploit spectral energy concentration.
Piao et al.~\cite{fredformer} identify a low-frequency bias in Transformers, which favors low-frequency features over high-frequency components.
They propose Fredformer~\cite{fredformer}, which processes spectral patches with independent Transformer blocks to mitigate this bias.
\end{itemize}

\subsection{Plug-in Layers}\label{sec:plugin}

\begin{figure}
    \centering
    \includegraphics[width=\linewidth]{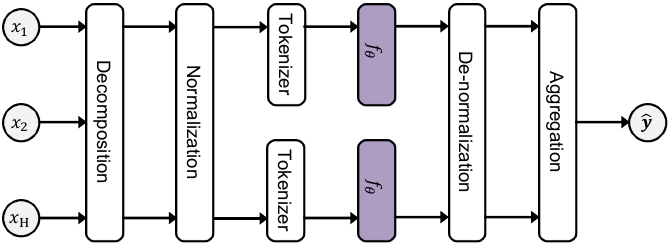}
    \caption{A typical layout of the plug-in layers following~\cite{Tempo}, where the colored blocks represent the backbone architecture.}
    \label{fig:plugin}
\end{figure}

Plug-in layers aim to help standard backbones model complex autocorrelation structures with minimal architectural modification. Exemplars include: normalization, decomposition, and tokenization layers. As shown in \autoref{fig:plugin}, they can be seamlessly integrated into a unified pipeline: the input series is first normalized and decomposed; the resulting components are then tokenized and processed independently by the backbone. Finally, their outputs are aggregated and denormalized to produce the predicted sequence.

\subsubsection{Normalization Layers}\label{sec:normalization}
The distribution statistics of $\mathbf{x}$, such as its mean and variance, may change over time because of non-stationarity \cite{zhou2018multimode}. For instance, a persistent upward trend produces an increasing mean~\cite{chen2023mode}.
These changes cause distribution shifts between training and test data, impairing generalization~\cite{Stationary,RevIN}.
To address this, normalization layers standardize the statistics of $\mathbf{x}$ across samples.
A common approach is Z-normalization~\cite{Stationary}.
Given a history sequence $\mathbf{x}$ with mean $\mu$ and standard deviation $\sigma$, normalization, forecasting, and denormalization proceed as:
\begin{equation*}
    \begin{aligned}
    \mathbf{x}^\mathrm{(n)} &= \frac{\mathbf{x} - \mu}{\sigma + \epsilon}, \qquad \hat{\mathbf{y}}^\mathrm{(n)} = f_\theta(\mathbf{x}^\mathrm{(n)}),\\
    \hat{\mathbf{y}} &= (\sigma+\epsilon)\hat{\mathbf{y}}^\mathrm{(n)} + \mu,
    \end{aligned}
\end{equation*}
where $\epsilon$ ensures numerical stability.
This procedure enforces zero mean and unit variance across input sequences.
RevIN~\cite{RevIN} extends this approach with learnable affine parameters for greater adaptability and is widely adopted in recent architectures~\cite{RLinear,Film,TSMixer2,SAMformer}.

However, Z-normalization and RevIN assume that $\mathbf{x}$ and $\mathbf{y}$ share identical statistics, which may not hold given non-stationarity.
To address this, prediction-enhanced normalization methods have been proposed~\cite{san,dish-ts,sin,fan,ddn}, which learn to predict the statistics of $\mathbf{y}$ for denormalization.
For instance, Dish-TS~\cite{dish-ts} employs learnable modules to estimate the mean and variance of $\mathbf{y}$.
Building on this, SAN~\cite{san} performs normalization and statistic prediction at the patch level rather than on the entire sequence  to accommodate localized distribution shifts.
Beyond traditional Z-normalization, SIN~\cite{sin} learns a fully adaptive normalization scheme to enhance adaptivity.

\subsubsection{Decomposition Layers}\label{sec:decomposition}
Multiple autocorrelation patterns often co-exist in $\mathbf{x}$ jointly (\eg  mixed trend and seasonality), and it is difficult to model them jointly.
To address this, decomposition layers decompose $\mathbf{x}$ into components with simpler autocorrelation structures that can be modeled by dedicated submodules~\cite{dudek2023std}.
Existing methods fall into three categories.
\begin{itemize}[leftmargin=*]
    \item \textbf{Seasonal-trend decomposition}: these methods extract a smooth trend component from $\mathbf{x}$ via moving average and treat the residual as the seasonal component:
    \begin{equation*}
        \begin{aligned}
            \mathbf{x}^\mathrm{(trend)}=\mathrm{MovingAvg}(\mathbf{x}), \quad \mathbf{x}^\mathrm{(season)}=\mathbf{x}-\mathbf{x}^\mathrm{(trend)},
        \end{aligned}
    \end{equation*}
    which is widespread in modern architectures~\cite{Autoformer,xPatch,patchmlp,Tempo,MICN}.
For example, Autoformer~\cite{Autoformer} performs this decomposition within each block and focuses representation learning on the seasonal component; xPatch~\cite{xPatch} performs this decomposition at the input stage and processes the two components using separate network branches.
    \item \textbf{Periodic decomposition}: these methods decompose $\mathbf{x}$ into segments according to dominant periods, which are identified from prominent peaks in the frequency spectrum~\cite{Times2d,Timesnet}.
    For a period $\mathrm{P_e}$, $\mathbf{x}$ is folded as:
    \begin{equation*}
    \mathbf{Z}^\mathrm{(\mathrm{P_e})} = \mathrm{Fold}(\mathbf{x}), \quad \mathbf{Z}^\mathrm{(\mathrm{P_e})} \in \mathbb{R}^{\frac{\mathrm{H}}{\mathrm{P_e}} \times \mathrm{P_e}},
    \end{equation*}
    where each row corresponds to one period cycle.
    An exemplar is TimesNet~\cite{Timesnet}, which decomposes the series into multiple periodic components and processes each independently with an Inception block.
    \item \textbf{Resolution decomposition}: these methods downsample $\mathbf{x}$ with different factors $\mathrm{P_r}$ to obtain components:
    \begin{equation*}
        \mathbf{x}^\mathrm{(\mathrm{P_r})}=\mathrm{Downsample}(\mathbf{x}, \mathrm{P_r}),
    \end{equation*}
    which reflect autocorrelations at different granularities, supporting multiscale architectures~\cite{SCINet,wang2024timemixer,wang2025timemixer++}.
For example, TimeMixer~\cite{wang2024timemixer} uses average pooling to construct multi-resolution components, processes them with MLPs, and aggregates scale-specific outputs to produce the final forecast.
\end{itemize}

\subsubsection{Tokenization Layers}\label{sec:tokenization}
Individual observations in $\mathbf{x}$ carry limited semantic information without temporal context, as discussed in Section~\ref{sec:standardselfattention}.
To address this, tokenization layers transform $\mathbf{x}$ into a series of context-rich tokens~\cite{timesql,patchmlp}.
A common approach is patch-based tokenization, popularized by PatchTST~\cite{PatchTST}, which segments $\mathbf{x}$ into patches using a sliding window of length $\mathrm{P}_{\mathrm{w}}$ and stride $\mathrm{P}_{\mathrm{s}}$:
\begin{equation*}
    \begin{aligned}
        \mathbf{P}&=\mathrm{SlideWindow}(\mathbf{x}),\\
        \mathbf{p}_h &= [x_{1+(h-1)\mathrm{P}_\mathrm{s}}, \ldots, x_{(h-1)\mathrm{P}_\mathrm{s}+\mathrm{P}_\mathrm{w}}],
    \end{aligned}
\end{equation*}
where $\mathbf{P}\in\mathbb{R}^{\mathrm{N}\times\mathrm{P}_{\mathrm{w}}}$ stacks the patches as rows, $\mathbf{p}_h$ denotes the $h$-th patch, and $\mathrm{N}=\lfloor(\mathrm{H}-\mathrm{P}_{\mathrm{w}})/\mathrm{P}_{\mathrm{s}}\rfloor+1$ without padding.
The obtained patches are typically linearly projected to match the standard backbone's input dimension~\cite{hucontext,tan2024language}.
Such patch-based tokenizer is widely adopted in modern architectures, especially Transformer-based models~\cite{Time-LLM,GPT4TS}.
However, not all patches benefit forecasting equally, which motivates studies to identify and discard less informative patches ~\cite{wusrsnet}.

The autocorrelation in real-world time-series often exhibits multiscale characteristics (\eg short-term seasonality and long-term cyclicity), which has catalyzed the development of multiscale tokenizers~\cite{timesql,patchmlp}.
For example, ElasTST~\cite{ElasTST} and MTST~\cite{MTST} apply multiple sliding windows with distinct $\mathrm{P}_\mathrm{w}$ and $\mathrm{P}_\mathrm{s}$ to capture autocorrelation across different scales.
The resulting token series are processed independently and subsequently aggregated for forecasting.
WaveToken~\cite{WaveToken} and SimpleTM~\cite{Simpletm} tokenize $\mathbf{x}$ via wavelet decomposition, which produces multiscale token series with little redundancy.
These strategies enhance the capacity to represent complex autocorrelations without major architectural changes to the core forecasting backbone.

\section{Loss Functions for Modeling Autocorrelation}\label{sec:loss}

Loss function is another critical ingredient for deep time-series forecasting.
The central challenge is to effectively model \textit{label autocorrelation}, \ie the autocorrelation within the label sequence.
As a critical property, neglecting label autocorrelation leads to suboptimal loss functions and degraded forecasting performance.

A standard loss function is the mean squared error (MSE):
\begin{equation}\label{eq:tmp}
\mathcal{E}_\mathrm{MSE}=\frac{1}{\mathrm{T}}\sum_{t=1}^\mathrm{T}\left(y_t-\hat{y}_t\right)^2
\end{equation}
which measures the pointwise squared error between $\mathbf{y}$ and $\hat{\mathbf{y}}$.
Despite its prevalence~\cite{FreTS,itransformer,OLinear}, $\mathcal{E}_\mathrm{MSE}$ has a critical limitation: it assumes that different elements in $\mathbf{y}$ are conditionally uncorrelated given $\mathbf{x}$.
Such assumption disregards label autocorrelation structure where $y_t$ depends on its predecessors $y_{<t}$~\citep{DLinear}.
To address this limitation of $\mathcal{E}_\mathrm{MSE}$, various loss functions are developed, which primarily fall into four groups: likelihood estimation, shape alignment, distribution balancing, and conditional generation.

\begin{table*}
    \centering
    \caption{Overview of loss functions harnessing label autocorrelation.}
    \label{tab:obj}
    \scriptsize
    \renewcommand{\arraystretch}{0.9} 
    \begin{tabular}{
        >{\raggedright\arraybackslash}p{2cm} 
        >{\raggedright\arraybackslash}p{1.5cm}   
        >{\raggedright\arraybackslash}p{2.9cm} 
        >{\centering\arraybackslash}p{1.5cm}   
        >{\centering\arraybackslash}p{1.5cm} 
        >{\centering\arraybackslash}p{1.5cm}    
        >{\centering\arraybackslash}p{1.5cm} 
        >{\centering\arraybackslash}p{1.5cm} 
    }
    \toprule
    \textbf{Loss function} & \textbf{Year} & \textbf{Type}  & \textbf{Agnostic} & \textbf{AutoDiff} & \textbf{Analytical} & \textbf{Infer. free} & \textbf{Param. free} \\
    \midrule
    \rowcolor{lightgray} MSE~\cite{Informer} & 2021 & \false  & \true & \true & \true & \true & \true \\
    AutoMSE~\cite{AutoMSE} & 2021 & Likelihood estimation  & \true & \true & \true & \true & \true \\
    \rowcolor{lightgray} FreDF~\cite{wang2025iclrfredf} & 2025 & Likelihood estimation  & \true & \true & \true & \true & \true \\
    Time-o1~\cite{wang2025nipstimeo1} & 2025 & Likelihood estimation  & \true & \true & \true & \true & \true \\
    \rowcolor{lightgray} LatentTSF~\cite{yang2026observations} & 2026 & Likelihood estimation  & \true & \true & \true & \false & \false \\
    DBLoss~\cite{qiudbloss} & 2025 & Likelihood estimation  & \true & \true & \true & \true & \true \\
    \rowcolor{lightgray} TDAlign~\cite{tdalign} & 2025 & Likelihood estimation  & \true & \true & \true & \true & \true \\
    QDF~\cite{wang2026iclrqdf} & 2026 & Likelihood estimation  & \true & \true & \false & \false & \false \\
    \rowcolor{lightgray} DTW~\cite{dtw} & 2003 & Shape alignment  & \true & \false & \false & \true & \true \\
    SoftDTW~\cite{soft-dtw} & 2017 & Shape alignment & \true & \true & \false & \true & \true \\
    \rowcolor{lightgray} Dilate~\cite{Dilate} & 2019 & Shape alignment  & \true & \true & \false & \true & \true \\
    GromovDTW~\cite{GDTW2} & 2021 & Shape alignment & \true & \true & \false & \true & \true \\
    \rowcolor{lightgray} AST~\cite{AST} & 2020 & Distribution balancing  & \true & \true & \false & \true & \false \\
    SDGCN~\cite{SDGCN} & 2025 & Distribution balancing  & \true & \true & \false & \true & \false \\
    \rowcolor{lightgray} AttnWGAIN~\cite{AttnWGAIN} & 2025 & Distribution balancing  & \true & \true & \false & \true & \false \\
    PSLoss~\cite{psloss} & 2025 & Distribution balancing  & \true & \true & \true & \true & \true \\
    \rowcolor{lightgray} DistDF~\cite{wang2026iclrdistdf} & 2026 & Distribution balancing  & \true & \true & \true & \true & \true \\
    TimeGrad~\cite{TimeGrad} & 2021 & Conditional generation  & \true & \true & \true & \false & \true \\
    \rowcolor{lightgray} CSDI~\cite{CSDI} & 2021 & Conditional generation  & \true & \true & \true & \false & \true \\
    D3M~\cite{D3M} & 2024 & Conditional generation  & \true & \true & \true & \false & \true \\
    \rowcolor{lightgray} AutoTimes~\cite{Autotimes} & 2024 & Conditional generation  & \true & \true & \true & \false & \true \\
    Timer~\cite{Timer} & 2024 & Conditional generation  & \true & \true & \true & \false & \true \\
    \rowcolor{lightgray} Timer-XL~\cite{Timer-XL} & 2025 & Conditional generation  & \true & \true & \true & \false & \true \\
    MoLA~\cite{MoLA} & 2025 & Conditional generation  & \true & \true & \true & \false & \true \\
    \bottomrule
    \end{tabular}
    \begin{tablenotes}
        \item \textit{Note:} ``Agnostic'' indicates whether the loss function is agnostic to backbone architectures, ``AutoDiff'' indicates whether it permits automatic differentiation, ``Analytical'' indicates whether it admits a closed-form solution without numerical approximation, ``Infer. free'' indicates whether it does not add inference time,  ``Param. free'' indicates whether it introduces no learnable parameters.
    \end{tablenotes}
    \end{table*}

\subsection{Likelihood Estimation}\label{sec:likelihoodestimation}

\begin{figure}
    \centering
    \subfigure[Label transformation.]{\includegraphics[width=0.48\columnwidth]{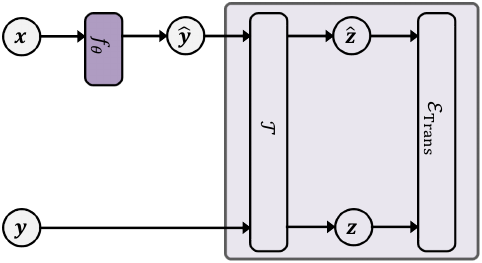}}\hfill
    \subfigure[Covariance modeling.]{\includegraphics[width=0.48\columnwidth]{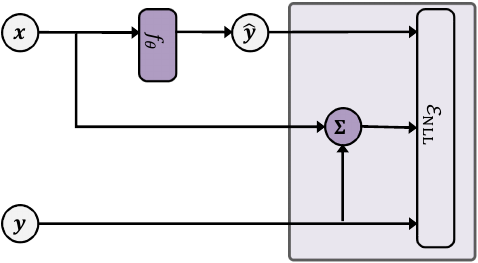}}
    \caption{The computation of loss functions based on likelihood estimation, where the dark blocks contain learnable parameters.}
    \label{fig:obj1}
\end{figure}

Likelihood estimation methods leverage the negative log-likelihood (NLL) to derive loss functions.
Under a Gaussian assumption, the NLL of $\mathbf{y}$ is roughly given by~\cite{wang2025nipstimeo1}:
\begin{equation}\label{eq:nll}
    \begin{aligned}
        \mathcal{E}_\mathrm{NLL} 
        &= (\mathbf{y}-\hat{\mathbf{y}})^\top\mathbf{\Sigma}^{-1}(\mathbf{y}-\hat{\mathbf{y}})+\log\left|\mathbf{\Sigma}\right|\\
        &=\left\|\mathbf{y}-\hat{\mathbf{y}}\right\|_{\mathbf{\Sigma}^{-1}}^2+\log\left|\mathbf{\Sigma}\right|,
    \end{aligned}
\end{equation}
where $\mathbf{\Sigma}$ is the conditional covariance matrix of $\mathbf{y}$.
$\mathcal{E}_\mathrm{MSE}$ in~\eqref{eq:tmp} is a special case of $\mathcal{E}_\mathrm{NLL}$ that assumes $\boldsymbol{\Sigma} = \mathbf{I}$.
This assumption ignores label autocorrelation, creating a bias between $\mathcal{E}_\mathrm{MSE}$ and the true $\mathcal{E}_\mathrm{NLL}$.
This issue is termed \textit{autocorrelation bias} and formalized in Theorem~\ref{thm:bias}.
Existing methods address this bias through two main strategies (see \autoref{fig:obj1}): transforming $\mathbf{y}$ into uncorrelated latent components or directly learning $\boldsymbol{\Sigma}$.

\begin{theorem}[Autocorrelation bias~\cite{wang2025nipstimeo1}]
    \label{thm:bias}
    Let $\mathbf{y}\in\mathbb{R}^\mathrm{T}$ be a label sequence with conditional covariance matrix $\mathbf{\Sigma}\in\mathbb{R}^{\mathrm{T}\times\mathrm{T}}$.
The $\mathcal{E}_\mathrm{MSE}$ in~\eqref{eq:tmp} is biased against $\mathcal{E}_\mathrm{NLL}$, expressed as:
    \begin{equation}
        \begin{aligned}
            \mathrm{Bias} 
            &= \mathcal{E}_\mathrm{NLL}-\mathcal{E}_\mathrm{MSE}\\
            &=\left\|\mathbf{y}-\hat{\mathbf{y}}\right\|_{\mathbf{\Sigma}^{-1}}^2 - \left\|\mathbf{y}-\hat{\mathbf{y}}\right\|^2 -\frac{1}{2}\log\left|\mathbf{\Sigma}\right|.
        \end{aligned}
    \end{equation}
    This bias vanishes if and only if $\mathbf{\Sigma}$ is identity, \ie different steps in $\mathbf{y}$ are conditionally decorrelated.
\end{theorem}

\subsubsection{Label Transformation}\label{sec:labeltransform}
These methods employ a mapping function $\mathbb{T}$ to transform $\mathbf{y}$ into latent components $\mathbf{z}$.
Given $p\geq 1$, the loss function is defined as:
\begin{equation}\label{eq:trans}
\mathcal{E}_\mathrm{Trans} = \left\|\mathbb{T}(\mathbf{y})-\mathbb{T}(\mathbf{\hat{y}})\right\|_p^p:=\left\|\mathbf{z}-\mathbf{\hat{z}}\right\|_p^p.
\end{equation}
The premise is that $\mathbf{z}$ possesses a simplified autocorrelation structure, rendering it more amenable to pointwise loss like $\mathcal{E}_\mathrm{MSE}$.
Methods primarily differ in their choice of $\mathbb{T}$:
\begin{itemize}[leftmargin=*]
    \item \textbf{Label whitening methods}: they employ \textit{whitening functions} as $\mathbb{T}$ to generate decorrelated components, thereby diminishing autocorrelation bias (Theorem~\ref{thm:bias}).
FreDF~\cite{wang2025iclrfredf} instantiates $\mathbb{T}$ as the normalized discrete Fourier transform (DFT), yielding the loss function:
    \begin{equation*}
        \mathcal{E}_\mathrm{FreDF} = \left\|\mathrm{DFT}(\mathbf{y})-\mathrm{DFT}(\mathbf{\hat{y}})\right\|_p^p,
    \end{equation*}
    which is straightforward to implement and has served as a foundational baseline in many studies~\cite{fredf_app1,fredf_app2,fredf_app3}.
    However, the normalized DFT is only asymptotically whitening (see Theorem 3.3 in \cite{wang2025iclrfredf}) and thus does not guarantee decorrelated components when $\mathbf{y}$ has finite length.
    To address this limitation, Time-o1~\cite{wang2025nipstimeo1} instantiates $\mathbb{T}$ using principal component analysis, yielding decorrelated components when $\mathbf{y}$ has finite length.
    However, it requires singular value decomposition over the entire training set to compute $\mathbb{T}$, which can incur substantial cost for exceptionally large-scale datasets. More recently, LatentTSF~\cite{yang2026observations} further learns adaptive $\mathbb{T}$ with autoencoding, tailoring it to dataset-specific characteristics and improving forecasting performance across benchmarks. Notably, theoretical results in \cite{wang2025iclrfredf,wang2025nipstimeo1} only guarantee marginal decorrelation of obtained components, not the conditional decorrelation required by Theorem \ref{thm:bias}. Nevertheless, empirical studies show the obtained components are often conditionally decorrelated \cite{wang2025nipstimeo1}.
    \item \textbf{Label stabilization methods:} they use \textit{filtering functions} as $\mathbb{T}$ to reduce certain forms of nonstationarity or simplify local dynamics.
While these methods do not eliminate autocorrelation bias, they improve forecasting performance.
For example, TDAlign~\cite{tdalign} defines $\mathbb{T}$ as a forward difference operator ($z_t = \Delta y_t$).
AutoMSE~\cite{AutoMSE} computes residuals based on first-order autocorrelation ($z_t = y_{t+1}-\hat{\rho} y_t$), which may eliminate first-order dependencies given an accurate estimate of the autocorrelation coefficient ($\hat{\rho}$). 
    DBLoss~\cite{qiudbloss} separately models trend and seasonal components, leveraging seasonal stability to improve forecasting performance.
\end{itemize}

\subsubsection{Covariance Modeling}\label{sec:covariancemodeling}
These methods parameterize the covariance matrix $\mathbf{\Sigma}$ to model label autocorrelation, which is subsequently used to compute the NLL in~\eqref{eq:nll}.
They differ mainly in how $\boldsymbol{\Sigma}$ is estimated.
For example, MKE~\cite{MKE} constructs $\boldsymbol{\Sigma}$ as a linear combination of predefined kernel matrices, $\boldsymbol{\Sigma} = \sum_{i=1}^{\mathrm{K}} w_i \mathbf{K}_i$, where the weights $[w_1,...,w_\mathrm{K}]$ are dynamically predicted by an auxiliary network conditioned on the backbone architecture's hidden states.
Such an input-dependent formulation allows MKE to model conditional correlations dynamically.
The auxiliary network is jointly optimized with the backbone architecture to minimize the NLL.
MMKE~\cite{MMKE} extends MKE to multivariate forecasting by jointly modeling the temporal and cross-variate correlation.
More recently, QDF~\cite{wang2026iclrqdf} reframes covariance estimation as a meta-learning problem, treating $\boldsymbol{\Sigma}$ as meta-parameters to be optimized.
It learns $\boldsymbol{\Sigma}$ via bi-level optimization, alternating between several inner-loop updates of $f_\theta$ and outer-loop updates of $\boldsymbol{\Sigma}$.
Although QDF effectively improves forecasting performance, the bi-level optimization process incurs additional computational overhead.

\subsection{Shape Alignment}\label{sec:shapealignment}
\begin{figure}
    \centering
    \includegraphics[width=0.98\columnwidth]{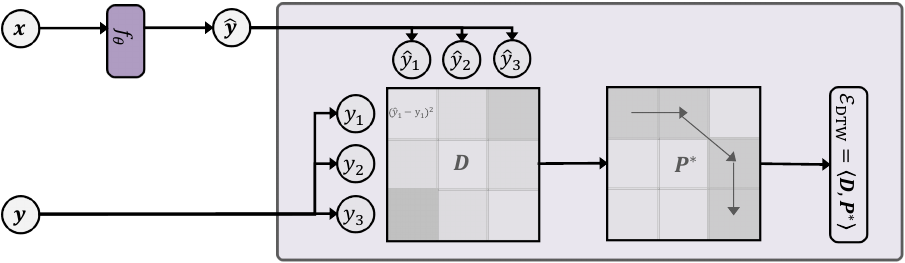}
    \caption{The computation of loss functions based on shape alignment, where the dark blocks contain learnable parameters.}
    \label{fig:obj2}
\end{figure}

Shape alignment methods leverage shape dissimilarity to define loss functions.
The premise is that morphology encodes autocorrelation structure, so morphological comparison yields autocorrelation-aware loss.
To this end, a classical approach is dynamic time warping (DTW), which treats $\mathbf{y}$ and $\hat{\mathbf{y}}$ as sandpiles of different shapes and defines their dissimilarity as the minimal cost to reshape one into the other~\cite{chen2013dtw}. A visualization is available in \autoref{fig:obj2}.

\begin{definition}\label{def:dtw}
\textit{Given $\mathbf{y} = [y_1, \ldots, y_\mathrm{T}]$ and $\hat{\mathbf{y}} = [\hat{y}_1, \ldots, \hat{y}_\mathrm{T}]$, the DTW distance between them is defined as the solution to the following constrained optimization problem:}
\begin{equation*}\label{eq:dtw}
\mathcal{E}_\mathrm{DTW}(\mathbf{y}, \hat{\mathbf{y}})
:= \min_{\mathbf{P} \in \Pi(\mathrm{T}, \mathrm{T})}
\langle \mathbf{C}, \mathbf{P} \rangle,
\end{equation*}
\textit{where $\boldsymbol{C}$ is the pairwise cost matrix with entries $c_{ij} = (y_i - \hat{y}_j)^2$, $\langle \cdot, \cdot \rangle$ denotes the Frobenius inner product, and $\Pi(m, n)$ is the set of all admissible alignment matrices, where an admissible matrix $\mathbf{P}$ defines a monotonic and continuous warping path from $(1,1)$ to $(m,n)$ via steps of $\downarrow$, $\rightarrow$, or $\searrow$.}
\end{definition}

By Definition~\ref{def:dtw}, DTW computes the optimal alignment matrix $\mathbf{P}$, whose elements $p_{i,j}$ depend recursively on predecessors ($p_{i-1,j}$ and $p_{i,j-1}$), thereby accommodating autocorrelation within $\mathbf{y}$.
However, its non-differentiability precludes direct use as a loss in training neural networks.
To address this, SoftDTW~\cite{soft-dtw} introduces a differentiable relaxation:
\begin{equation}\label{eq:sdtw}
    \mathcal{E}_\mathrm{Soft}(\mathbf{y}, \hat{\mathbf{y}}) 
    := -\gamma \log\sum_{\mathbf{P} \in \Pi(m, n)} \exp(-\langle \mathbf{C}, \mathbf{P} \rangle/\gamma),
\end{equation}
with smoothing temperature $\gamma > 0$.
It has been a prevalent loss function across domains, such as speech processing~\cite{zeitler2026unified}, healthcare~\cite{softdtw_health2}, and geoscience~\cite{softdtw_geoscience}. 

Subsequent work extends SoftDTW to address its limitations.
DSDTW~\cite{DSDTW} notes that SoftDTW can yield negative values and develops a divergence-based variant with positivity guarantees. 
Dilate~\cite{Dilate} fuses SoftDTW with a temporal distortion index, \ie the area between the optimal alignment matrix and the first diagonal, to quantify and reduce lag mismatches. 
STRIPE~\cite{STRIPE2} builds upon Dilate for probabilistic forecasting to promote structured diversity in predictions.
More innovations include using a Gumbel-Softmin operator for linear complexity~\cite{GDTW}, limiting warping path lenght to eliminate spurious alignments~\cite{LDTW}, using Gromov-Wasserstein distance to compare series from incomparable spaces~\cite{GDTW2}, and promoting matching steps with similar neighbors~\cite{ShapeDTW}.

\subsection{Distribution Balancing}\label{sec:distributionbalancing}

\begin{figure}
    \centering
    \subfigure[Discrepancy minimization.]{\includegraphics[width=0.48\columnwidth]{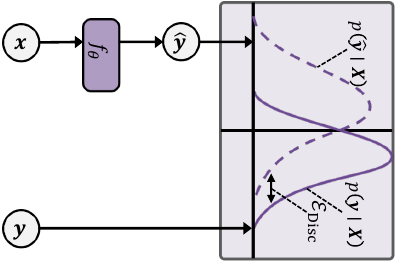}}\hfill
    \subfigure[Adversarial training.]{\includegraphics[width=0.48\columnwidth]{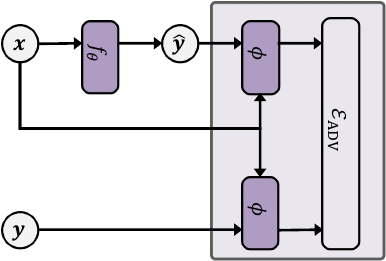}}
    \caption{The computation of loss functions based on distribution balancing, where the dark blocks contain learnable parameters.}
    \label{fig:obj3}
\end{figure}

Distribution balancing methods treat time-series forecasting as a conditional distribution alignment problem, and define the loss function as the discrepancy between the conditional distribution of $\mathbf{y}$ and $\hat{\mathbf{y}}$:
\begin{equation}\label{eq:discrepancy}
    \mathcal{E}_\mathrm{Disc} = \mathrm{Disc}(p(\mathbf{y}| \mathbf{x}), p(\hat{\mathbf{y}}| \mathbf{x})),
\end{equation}
where $\mathrm{Disc}$ represents distributional discrepancy.
By the property of ideal distributional discrepancy, the discrepancy vanishes iff the two distributions coincide. Thus, minimizing $\mathcal{E}_\mathrm{Disc}$ can align the full conditional distributions, including their label autocorrelation structures.
However, designing a discrepancy well-suited to time-series data remains challenging, and current methods differ primarily in how they define it (see \autoref{fig:obj3}).

\subsubsection{Discrepancy Minimization}\label{sec:discrepancyminimization}
These methods explicitly define and minimize $\mathrm{Disc}$.
An early approach, PSLoss~\cite{psloss}, defines $\mathrm{Disc}$ using first- and second-order moments.
Although computationally efficient, such low-order statistics may not fully characterize complex label distributions, leaving residual misalignment.
DistDF~\cite{wang2026iclrdistdf} implements $\mathrm{Disc}$ with Wasserstein distance, which can, in principle, compare full distributional characteristics beyond low-order moments~\cite{wang2025kddcfrpro,wang2025toiswbm,wang2023nipsescfr}. However, computing Wasserstein distance requires solving a constrained optimal transport problem, which is costly for long sequences.
To reduce this cost, DistDF uses the Bures-Wasserstein distance as a closed-form efficient approximation. Alternatively, KMB-DF~\cite{pan2026kmbdf} defines $\mathrm{Disc}$ in a reproducing kernel Hilbert space, enabling efficient comparison of distributional moments of all orders.

These advances motivate two further research directions.
The first line develops discrepancy tailored to time-series, as generic discrepancies may inadequately accommodate autocorrelation structure.
For example, PSW~\cite{wang2025iclrpswi} augments the Wasserstein distance with a frequency-domain pairwise distance to capture autocorrelation and a selective matching regularizer to improve noise robustness.
The second line introduces distribution alignment between $\mathbf{x}$ and $\mathbf{y}$ as a regularization loss, motivated by their shared distributional characteristics under appropriate stationarity assumptions.
To this end, TimeAlign~\cite{hu2026bridging} learns label representations of $\mathbf{y}$ via a reconstruction task, and aligns them with predictive representations of $\mathbf{x}$ through patch-wise and global relational losses, which improves forecasting performance across benchmarks.

\subsubsection{Adversarial Training}\label{sec:adversarialbalancing}
Because explicitly defining a $\mathrm{Disc}$ well-suited to time-series is difficult, these methods define it implicitly via adversarial optimization~\cite{goodfellowGAN}.
Specifically, they employ a mapping $\phi$ to transform $\mathbf{y}$ and $\hat{\mathbf{y}}$ into a latent space, and then quantify $\mathrm{Disc}$ as how readily their conditional distributions in the latent space can be distinguished given $\mathbf{x}$.
A large $\mathrm{Disc}$ implies greater distinguishability.
Since $\phi$ is unspecified, a natural choice is the worst-case discrepancy, obtained by maximizing over all possible $\phi$:
\begin{equation}\label{eq:discriminator}
    \begin{aligned}
    \phi^*&=\arg \max_{\phi\in\Phi} \mathrm{Disc}\left(p\left(\phi(\hat{\mathbf{y}})| \mathbf{x}\right), p\left(\phi(\mathbf{y})| \mathbf{x}\right)\right), \\
    \mathcal{E}_\mathrm{Adv} &= \mathrm{Disc}(p(\phi^*(\hat{\mathbf{y}})| \mathbf{x}), p(\phi^*(\mathbf{y})| \mathbf{x})),
    \end{aligned}
\end{equation}
which yields the adversarial discrepancy $\mathcal{E}_\mathrm{Adv}$.
Minimizing $\mathcal{E}_\mathrm{Adv}$ ensures that $\phi(\mathbf{y})$ and $\phi(\hat{\mathbf{y}})$ are indistinguishable even under the most discriminative mapping $\phi^*$.

An early representative method is AST~\cite{AST}, which adapts GAN~\cite{goodfellowGAN} to estimate and minimize $\mathcal{E}_\mathrm{Adv}$.
In this setup, $\phi$ acts as a discriminator (instantiated as an MLP), while the backbone $f_\theta$ acts as a generator.
Training alternates between two steps.
First, the discriminator $\phi$ is optimized via cross-entropy loss to distinguish between $\mathbf{y}$ and $\hat{\mathbf{y}}$, following~\eqref{eq:discriminator}.
The updated discriminator, denoted as $\phi^*$, provides an estimate of $\mathcal{E}_{\mathrm{Adv}}$.
Second, the generator $f_\theta$ is optimized to minimize $\mathcal{E}_\mathrm{Adv}$, aiming to produce predictions that are indistinguishable from $\mathbf{y}$. 

Building on AST, one line of work improves the backbone architecture of discriminator.
WRCGAN~\cite{WRCGAN} and SDGCN~\cite{SDGCN} replace the MLP discriminator with LSTM and GRU networks, respectively.
TrendGCN~\cite{TrendGCN} introduces an auxiliary discriminator that exploits sequence correlations.
Another line of work refines the formulation of adversarial loss.
For example, AttnWGAIN~\cite{AttnWGAIN} and WRCGAN~\cite{WRCGAN} adopt the Wasserstein GAN loss~\cite{WGAN} to improve training stability.
A common strategy among these works is to concatenate $\mathbf{x}$ with $\mathbf{y}$ and $\hat{\mathbf{y}}$ as the input to discriminator (see \autoref{fig:obj3}).
This strategy indeed aligns the joint distributions $p(\mathbf{x},\mathbf{y})$ and $p(\mathbf{x},\hat{\mathbf{y}})$; since both share the same marginal $p(\mathbf{x})$, matching these joint distributions is equivalent to matching the corresponding conditional distributions in~\eqref{eq:discrepancy} (see Lemma~3.3 in~\cite{wang2026iclrdistdf}).

\subsection{Conditional Generation}\label{sec:conditionalgeneration}

\begin{figure}
    \centering
    \subfigure[Diffusive generation.]{\includegraphics[width=0.50\columnwidth]{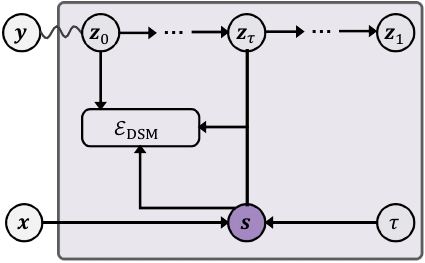}}\hfill
    \subfigure[Autoregressive generation.]{\includegraphics[width=0.46\columnwidth]{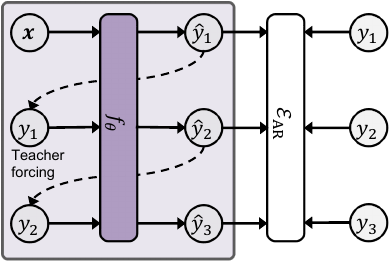}}
    \caption{The computation of loss functions based on conditional generation, where the dark blocks contain learnable parameters.}
    \label{fig:obj4}
\end{figure}

Conditional generation methods reframe time-series forecasting as a conditional generation problem: to producing label sequences conditioned on history sequences.
The premise is that generative models excel at modeling complex correlations within $\mathbf{y}$ for realistic synthesis~\cite{songscore}.
These methods exploit this property to accommodate label autocorrelation, which may enhance forecasting performance given ample data.
However, they also inherit the drawbacks of generative models, such as error propagation in autoregressive models and the high inference cost of diffusion models~\citep{TimeDiff}, which can affect their practical utility.
Existing work mainly explores two generative models: diffusion models and autoregressive models in \autoref{fig:obj4}.

\subsubsection{Diffusive Generation}\label{sec:diffusion}
Diffusion models have achieved strong results in image generation, which learns data distribution through an iterative denoising process~\cite{songscore}.
They are adapted for forecasting by treating $\mathbf{y}$ as a temporal image to be generated conditioned on $\mathbf{x}$~\cite{TimeDiff,StochDif}.
The target thus shifts from minimizing a deterministic pointwise error in \eqref{eq:tmp} to learning the conditional distribution $p(\mathbf{y}| \mathbf{x})$~\cite{D3VAE}.

A standard diffusion model comprises a forward noising process and a learned reverse denoising process.
The forward process progressively perturbs $\mathbf{y}$ into Gaussian noise, governed by the Itô process~\cite{songscore}:
\begin{equation}\label{eq:forward}
    \begin{aligned}
        d\mathbf{z}_\tau &= v_\tau(\mathbf{z}_\tau)\,d\tau
        + g_\tau\,d\boldsymbol{\omega}_\tau,
        \quad \tau \in [0,1],\\
        \mathbf{z}_0 &= \mathbf{y},
        \quad p_1(\mathbf{z}|\mathbf{x}) \approx \mathcal{N}(\mathbf{0},\mathbf{I}),
    \end{aligned}
\end{equation}
where $\boldsymbol{\omega}$ is a Wiener process representing the noise term, $v_\tau$ and $g_\tau$ are pre-defined drift and diffusion terms.
The reverse process generates predictions by denoising Gaussian samples conditioned on $\mathbf{x}$.
By Anderson's theorem~\cite{anderson1982reverse}, it follows:
\begin{equation}\label{eq:reverse}
    d\mathbf{z}_\tau =
    \big[v_\tau(\mathbf{z}_\tau)
    - g_\tau^2 \nabla_{\mathbf{z}_\tau}
    \log p_\tau(\mathbf{z}_\tau|\mathbf{x})\big]\,d\tau
    + g_\tau\,d\bar{\boldsymbol{\omega}}_\tau,
\end{equation}
which is integrated from $\tau=1$ to $\tau=0$.
Here, $\bar{\boldsymbol{\omega}}_\tau$ is a backward-time Wiener process, and $\nabla_{\mathbf{z}_\tau}\log p_\tau(\mathbf{z}_\tau|\mathbf{x})$ is the conditional score.
Since this score is unknown, a neural network $\mathbf{s}(\mathbf{z}_\tau,\tau,\mathbf{x})$ is trained to estimate it, which is trained using denoising score matching loss~\cite{songscore}:
\begin{equation}\label{eq:dsm}
    \mathcal{E}_{\mathrm{DSM}}
    = \mathbb{E}_{\tau,\mathbf{x},\mathbf{z}_0,\mathbf{z}_\tau}
    \left[
    \left\|
    \mathbf{s}(\mathbf{z}_\tau,\tau,\mathbf{x})
    - \nabla_{\mathbf{z}_\tau}
    \log p_\tau(\mathbf{z}_\tau|\mathbf{z}_0)
    \right\|_2^2
    \right],
\end{equation}
where $\tau\sim\mathcal{U}(0,1)$, $(\mathbf{x},\mathbf{z}_0)$ is sampled from the data distribution with $\mathbf{z}_0=\mathbf{y}$, and $\mathbf{z}_\tau\sim p_\tau(\mathbf{z}_\tau|\mathbf{z}_0)$.
Notably, $p(\mathbf{z}_\tau|\mathbf{z}_0)$ has a closed form given certain conditions, \eg $v_\tau(\mathbf{z})$ is linear in $\mathbf{z}$.
Once trained, the score estimator enables sampling from $p(\mathbf{y}|\mathbf{x})$ through the reverse process in~\eqref{eq:reverse}. This process implicitly models label autocorrelation, as in each step the estimation of $\mathbf{z}_\tau$ is conditioned on all elements in the previous step. 

The application of diffusion models in time-series forecasting is catalyzed by TimeGrad~\cite{TimeGrad}.
Subsequent research evolve along two complementary trajectories.
\begin{itemize}[leftmargin=*]
    \item \textbf{Improving architecture of the score network.}
    Accurate score estimation requires backbone architectures that effectively integrate the latent variable $\mathbf{z}_\tau$ with the conditioning input $\mathbf{x}$ while capturing autocorrelation within both. 
    To this end, early approaches employ CNNs~\cite{TimeDiff,li2018diffusion} and RNNs~\cite{TimeGrad,SSSD,SSD-TS}. Following the success of DiT~\cite{dit}, Transformer-based architectures have also gained prominence~\cite{Diffusion-TS,TMDM,D3U,TimeDart}. For example, CSDI~\cite{CSDI} fuses $\mathbf{z}_\tau$ and $\mathbf{x}$ through a one-dimensional convolutional layer and applies a hybrid attention module to model autocorrelation; TimeWeaver~\cite{TimeWeaver} extends CSDI with a preprocessing layer that incorporates metadata as auxiliary conditioning information. D3M~\cite{D3M} combines state-space dynamics with attention to model complex temporal dependencies while improving computational efficiency on long sequences.
    \item \textbf{Improving formulation of the diffusion process.}
    These works tailor diffusion process to time-series characteristics.
    Key innovations include frequency-domain formulations that exploit spectral energy concentration~~\cite{crabbetime}; input-dependent noise schedules that accommodate non-stationarity~\cite{NSDiff}, and conditional forward processes with non-isotropic terminal distributions that simplify reverse-process learning~\cite{CNDiff}.
     MR-Diff~\cite{MR-Diff} exploits the multi-resolution structure of time-series~\cite{MG-TSD}, extracting fine-to-coarse components in the forward pass and reconstructing coarse-to-fine details in the reverse pass, which offers physically meaningful dense supervision at each denoising step.
    TCDM~\cite{TCDM} introduces temporally correlated noise to retain autocorrelation in the latent states $\mathbf{z}_\tau$ as diffusion progresses.
\end{itemize}

\subsubsection{Autoregressive Generation}\label{sec:autoregressivegeneration}

Autoregression is widely used for sequence generation, from time-series forecasting models~\cite{DeepAR} to modern LLMs~\cite{guo2025deepseek}.
During inference, it generates predictions step by step: each prediction $\hat{y}_{\tau}$ is conditioned on $\mathbf{x}$ and preceding predictions $\hat{\mathbf{y}}_{<\tau}$.
During training, a common strategy is teacher forcing: in each step, it conditions  prediction on the preceding label sequence $\mathbf{y}_{<\tau}$:
\begin{equation*}
    \mathcal{E}_\mathrm{AR} = \sum_{\tau=1}^{\T} (\hat{y}_{\tau}-y_{\tau})^2 = \sum_{\tau=1}^{\T} \left(f_\theta(\mathbf{x}, \mathbf{y}_{<\tau})- y_{\tau}\right)^2.
\end{equation*}
By replacing $\hat{\mathbf{y}}_{<\tau}$ with $\mathbf{y}_{<\tau}$ during training, teacher forcing generally improves training stability.
Besides, training one step at a time circumvents the complexity of modeling label autocorrelation.

Autoregressive models suffer from error accumulation and high inference cost, which can be more pronounced as forecast horizon increases.
Direct forecasting, which generates the entire sequence $\hat{\mathbf{y}}$ in a single forward pass, therefore remains prevalent, yet it struggles to account for label autocorrelation.
Patch-wise autoregressive models offer a compromise between autoregressive and direct forecasting.
By generating a contiguous patch at each decoding step, they shorten the autoregressive chain while retaining dependencies across prediction patches.
Representative models include Timer~\cite{Timer}, Timer-XL~\cite{Timer-XL}, and TimeBase~\cite{Timebase}.
MoLA~\cite{MoLA} further introduces a low-rank adaptation scheme to learn patch-specific representations.
Patch-wise autoregression has been used to adapt pretrained LLMs for time-series forecasting.
These methods project time-series patches into the LLM's token embedding space and fine-tune the model to generate future patches autoregressively.
As LLMs are pretrained autoregressively, fine-tuning them in the same manner may better align with the pretraining task, facilitating adaptation to forecasting~\cite{Autotimes,LangTime}.

\section{Future Directions}\label{sec:futureworks}
This section outlines potential research directions for improving autocorrelation modeling and adapting autocorrelation modeling techniques to related tasks, both of which may contribute to advancing time-series forecasting

\subsection{Modeling Autocorrelation to Scale Backbones}\label{sec:futureobj}
Large-scale foundation model for time-series forecasting has attracted growing interest~\cite{Timer,Sundial}, inspired by scaling laws observed in other domains~\cite{guo2025deepseek}.
This direction faces two key challenges.
First, a general-purpose corpus suitable for pretraining is lacking.
Existing benchmarks (\eg  ETT~\cite{Informer}, M4~\cite{makridakis2018m4}) are often domain-specific and exhibit heterogeneous temporal patterns, making it difficult to learn transferable representations.
Second, time-series tokenization remains insufficiently explored.
As discussed in Section~\ref{sec:plugin}, raw observations in time-series carry limited semantics~\cite{itransformer,Crossformer}, necessitating effective tokenization to provide context-rich inputs and unlock the potential of large-scale foundation models.

Autocorrelation offers a unified perspective to address these challenges.
First, autocorrelation may facilitate the generation of high-quality synthetic data, which mitigates data scarcity.
As detailed in Section \ref{sec:preliminaries}, autocorrelation illustrates canonical patterns like trends and seasonality.
Prototypical autocorrelation structures across different domains may guide synthetic data generation or augmentation of existing datasets, helping models learn domain-invariant temporal dynamics.
Second, autocorrelation may shape the improved design of tokenizers.
While current patch-based tokenizers~\citep{PatchTST,xPatch,patchmlp} effectively encode temporal contexts, they are still less suited for scenarios with long-term, multiscale, or non-stationary autocorrelation.
Designing tokenizers that explicitly account for such complex autocorrelation patterns could improve the performance of backbone architectures in real-world time-series datasets.

\subsection{Modeling Autocorrelation with Advanced Generative Models}

Conditional generative models have gained interest for time-series forecasting due to their capacity to capture complex dependencies within label sequences.
However, existing approaches inherit drawbacks from their underlying generative models, such as the training instability of diffusion models and the error accumulation characteristic of autoregressive models.
These issues may be further exacerbated in time-series forecasting due to the typically long forecast horizons.
Refining and adapting generative models to alleviate these limitations is therefore a critical direction.

As the quick evolution of generative modeling landscape, advanced generative models (\eg  flow matching \cite{Sundial,kollovieh2025flow}, stochastic interpolants \cite{chen2024probabilistic}, and Schrodinger bridges \cite{hamdouche2026nonparametric,chen2023provably}) have emerged. These models remain largely underexplored in forecasting, yet may offer beneficial inductive biases and enhance forecasting performance.
For instance, Sundial~\cite{Sundial} shows that conditional flow matching yields a more tractable loss~\cite{kolloviehflow} compared with existing diffusion models while achieving competitive forecasting accuracy.

\subsection{Modeling Autocorrelation with Real-world Challenges}
Real-world time-series pose significant challenges for autocorrelation modeling and forecasting, yet methods are predominantly developed and evaluated on curated benchmarks that may not reflect them.
Consequently, benchmark improvements may not translate into practical performance, underscoring the need for methods that explicitly address these challenges.

The first challenge is measurement noise, which can arise from sensor inaccuracies, recording errors, and environmental interference.
One solution is denoised learning, which seeks to estimate an unbiased loss from noisy data \cite{wang2026causalrm,pan2026selectiverm}. For example, loss correction enables unbiased estimation of clean-data loss under suitable noise assumptions \cite{liu2015classification}. Recent studies further find that fusing a few clean data can relax restrictive assumptions and improve practical applicability \cite{glc}. A key limitation is that most loss-correction methods are designed for categorical labels~\cite{liu2015classification}, which requires further refinement to accommodate continuous labels in time-series forecasting.

The second challenge is distribution shift, which can arise from mode switches and external condition changes.
Such shifts may alter autocorrelation structures, compromising the generalization of models trained on historical data.
One approach is to train forecasting models that are robust to distribution shifts.
Representative strategies include domain adaptation, which leverages target-domain samples during training~\cite{wdgrl}, and domain generalization, which seeks to train generalizable models without access to the target domain \cite{li2018deep}.
Another approach is test-time adaptation, which updates model parameters using incoming observations to accommodate distribution shifts \cite{kim2025battling}.
However, in online settings, distinguishing noise and outliers from genuine changes in temporal dynamics is difficult, potentially misleading parameter updates. Effective test-time adaptation therefore requires balancing responsiveness to evolving dynamics with robustness to noise and outliers.

The third challenge is incomplete observation, which could arise from irregular sampling~\cite{yang2025revisiting}, sensor failures~\cite{wang2025taselsptd}, and differing sampling rates across sensors~\cite{zheng2025dynamic}.
A straightforward approach is to impute missing values before forecasting~\cite{wang2024tiispoti,wang2026tnnlspoti}. Modern distribution balancing strategies \cite{wang2025iclrpswi,yang2026glocal} have effectively supported imputation by harnessing global temporal structure to inform missing observations.
However, time-series imputation remains difficult, and its errors can distort the temporal structure and propagate into subsequent forecasts. Recent studies instead train forecasting models directly on incomplete observations, demonstrating robustness across varying missing rates~\cite{yang2025revisiting,liu2026astgi}. If multirate observations are available, mixture distribution and state-space models can integrate observations across sampling rates, demonstrating performance gains in large-scale industrial applications~\cite{lyu2023multirate,zheng2025dynamic}.

\subsection{Extending Autocorrelation Modeling to Channel Dependency Modeling}\label{sec:futurechannel}
Channel dependency refers to relationships among different variables, which is a central problem in multivariate time-series forecasting~\cite{han2024channel,qiu2025comprehensive}.
Examples include interactions among meteorological variables, coupled electricity demands across regions, and interdependent traffic flows at connected locations.
Modeling these dependencies can provide predictive information beyond that captured by autocorrelation within individual variables \cite{qiudag}, thereby offering opportunities to further improve real-world forecasting performance.

The techniques for modeling autocorrelation can be adapted to channel dependency modeling.
On the backbone architecture side, self-attention can model autocorrelation by aggregating information across temporal tokens, as well as channel dependencies by aggregating information across variate tokens, as exemplified by iTransformer~\cite{itransformer}.
On the loss function side, the label transformation strategy in Section~\ref{sec:labeltransform} can likewise be applied to reduce channel dependency~\cite{wang2025iclrfredf}.
Moreover, developing backbone architectures and loss functions to jointly accommodate autocorrelation and channel dependencies constitutes a promising research direction.

\section{Conclusion}

This paper presents a comprehensive survey of deep time-series forecasting through the lens of autocorrelation modeling.
It first establishes the central role of autocorrelation in forecasting and identifies two core challenges: modeling autocorrelation in history sequences and modeling autocorrelation in label sequences.
It then systematically reviews recent advances in backbone architectures and loss functions designed to address these respective challenges.
Building on this analysis, the survey outlines potential opportunities to further tackle the two challenges and advance the field.
In contrast to prior surveys, this work demonstrates that viewing both backbone architectures and loss functions through an autocorrelation-centric lens clarifies the field’s evolution and open challenges.

\textbf{Limitations.}
Although autocorrelation is fundamental to time-series forecasting, it does not cover all advances in the field, which also involve feature engineering, exogenous information, etc.
Besides, some advances can be understood from multiple perspectives. Our discussion emphasizes their connections to autocorrelation modeling to provide a coherent and focused description.

\scriptsize{
\bibliographystyle{ieeetr}
\bibliography{bib/abbr,bib/main}}

\begin{IEEEbiography}[{\includegraphics[width=1in,height=1.25in,clip,keepaspectratio]{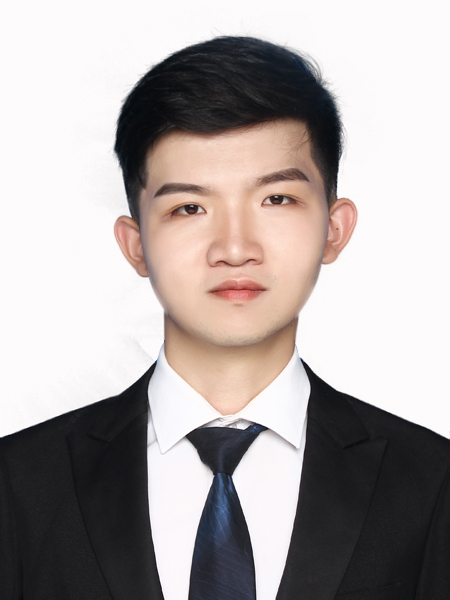}}]{Hao Wang}
    received the Ph.D. degree in control science from the College of Control Science and Engineering, Zhejiang University, Hangzhou, China. 
    His research interests include time-series analysis and causal machine learning.
    He has published more than 60 papers in top-tier AI venues such as ICML, NeurIPS, ICLR, and SIGKDD.
    He served as the PC member or Area Chair for top conferences such as ICML, NeurIPS, ICLR and SIGKDD, and the Associate Editor in IEEE SMC.
\end{IEEEbiography}

\begin{IEEEbiography}[{\includegraphics[width=1in,height=1.25in,clip,keepaspectratio]{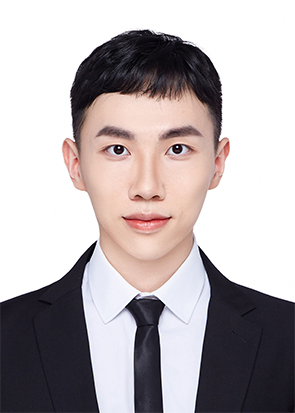}}]{Licheng Pan}
    received the B.Eng. degree in automation from the College of Control Science and Engineering, Zhejiang University, Hangzhou, China, in 2021. He is currently a Ph.D. student in cyberspace security with the College of Computer Science and Technology, Zhejiang University, Hangzhou, China.
His research interests include time-series analysis and trustworthy LLMs.
\end{IEEEbiography}

\begin{IEEEbiography}[{\includegraphics[width=1in,height=1.25in,clip,keepaspectratio]{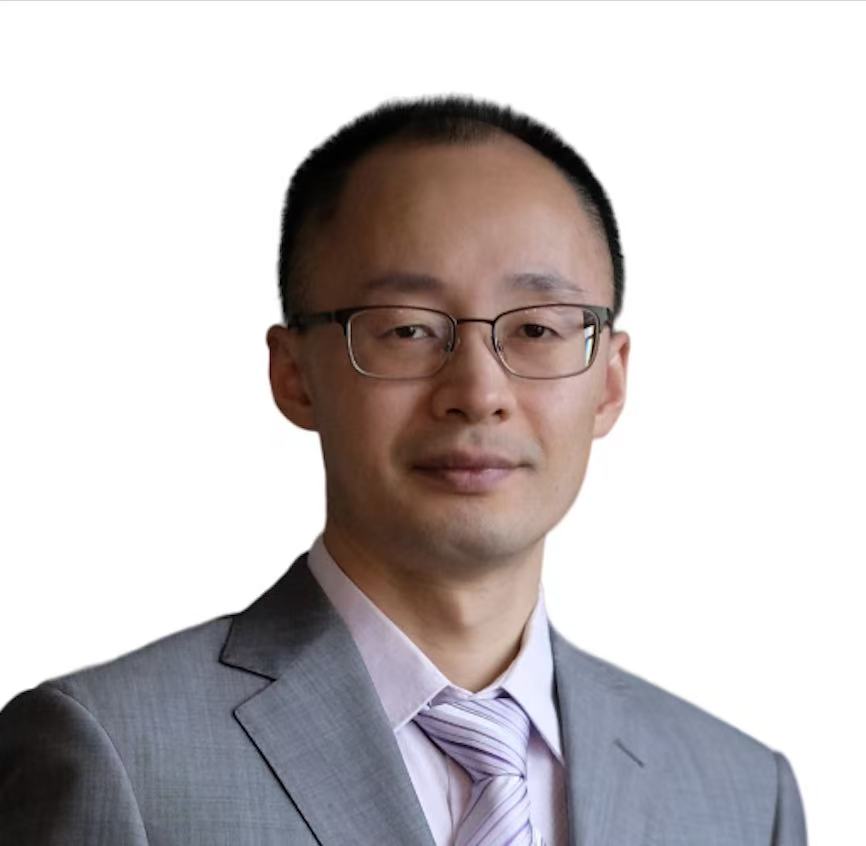}}]{Qingsong Wen} (Senior Member, IEEE) is currently the Head of AI \& Chief Scientist at Squirrel Ai Learning.
Before that, he worked at Alibaba, Qualcomm, Marvell, etc., and received his M.S. and Ph.D. degrees in Electrical and Computer Engineering from Georgia Institute of Technology, USA.
His research interests include machine learning, data mining, and signal processing.
He has published around 150 top-ranked AI conference and journal papers, had multiple Most Influential Papers at IJCAI, received multiple IAAI Innovative Application Awards at AAAI.
Currently, he serves as Chair of IEEE CIS Task Force on AI for Time-Series and Spatio-Temporal Data, and Vice Chair of INNS AI for Education Section.
He also regularly serves as Area Chair for top conferences including NeurIPS, ICML, KDD, IJCAI, etc, and Associate Editor for IEEE TPAMI and IEEE SPL.
\end{IEEEbiography}

\begin{IEEEbiography}[{\includegraphics[width=1in,height=1.25in,clip,keepaspectratio]{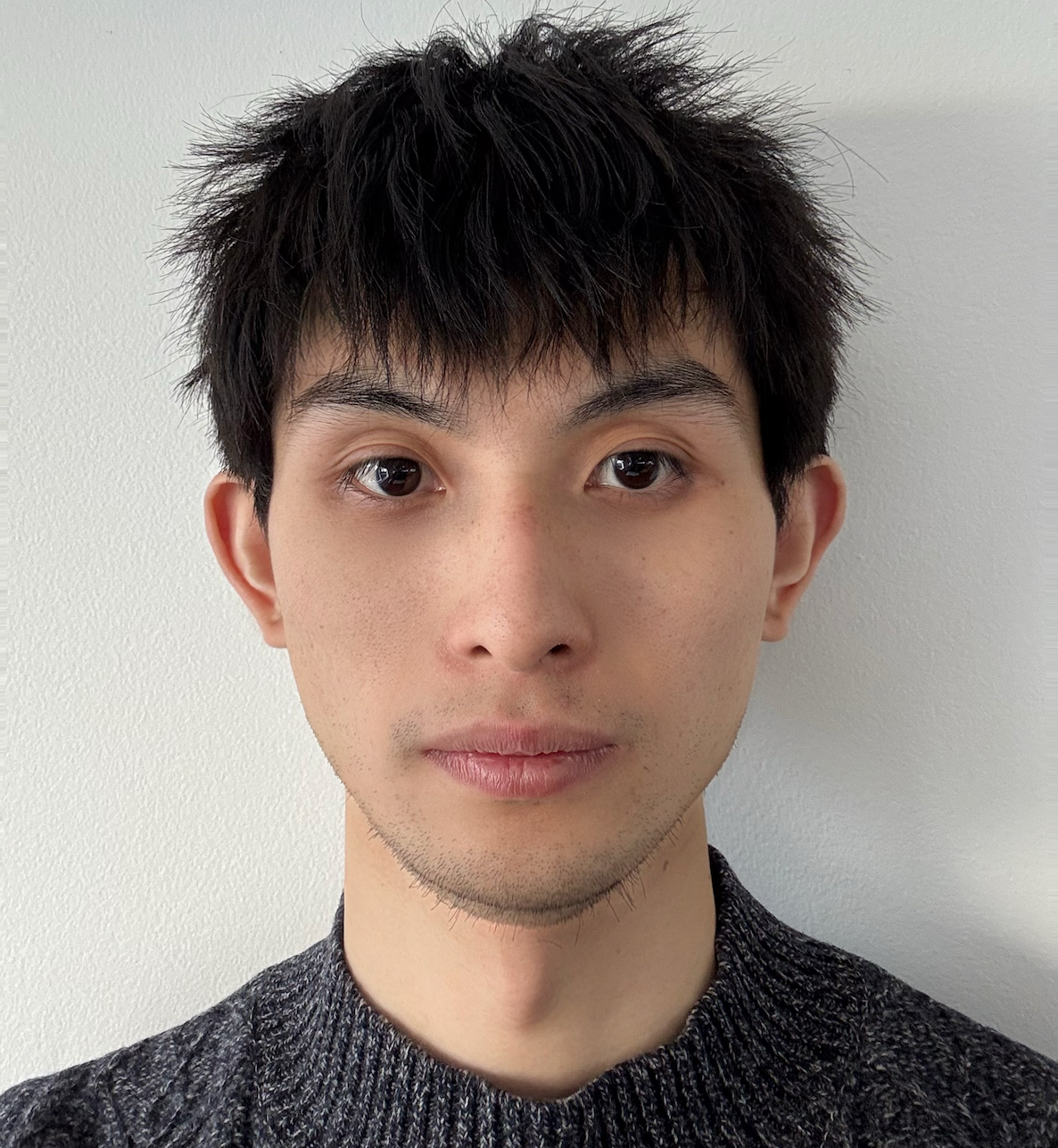}}]{Jialin Yu} is a Research Fellow at the University of Oxford and a Fellow of Christ Church, Oxford, and a Visiting Researcher at Microsoft.
He previously held research fellow positions at University College London and received his PhD from Durham University in 2023.
His research lies at the intersection of machine learning and causal inference, with the goal of developing AI systems that generalise reliably under distribution shift and novel deployment conditions.
\end{IEEEbiography}

\begin{IEEEbiography}[{\includegraphics[width=1in,height=1.25in,clip,keepaspectratio]{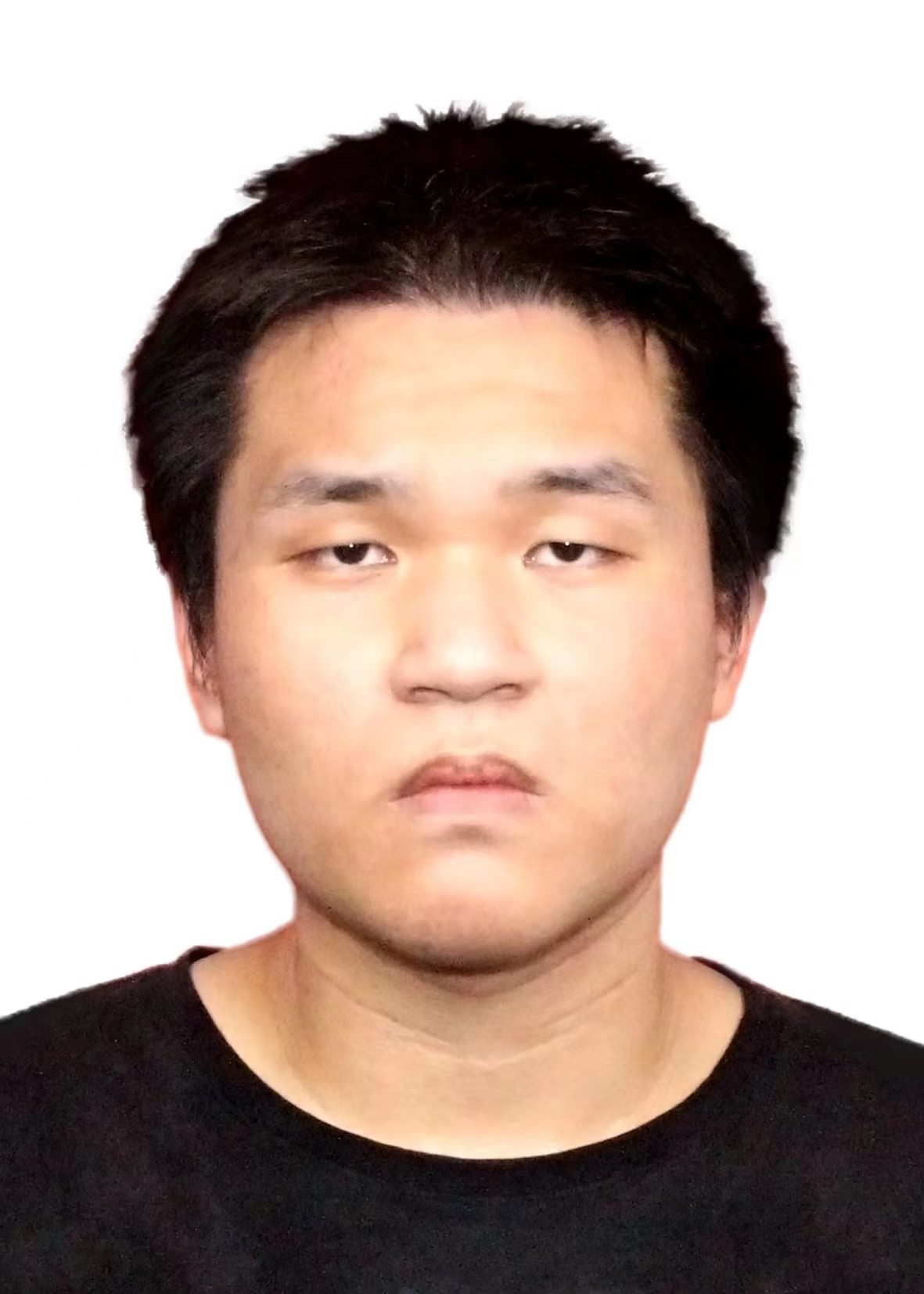}}]  {Zhichao Chen} received the Ph.D. degree from College of Control Science and Engineering, Zhejiang University, Zhejiang, China.
He is currently a post-doctoral research fellow at Peking University.
He has served as a reviewer for top-tier conferences including NeurIPS, ICML, ICLR, AAAI, and AISTATS.
He has also published papers at NeurIPS, ICML, ICLR, and IEEE TKDE.
His research interests include convex optimization, differential equation, and optimal control.
\end{IEEEbiography}

\begin{IEEEbiography}[{\includegraphics[width=1in,height=1.25in,clip,keepaspectratio]{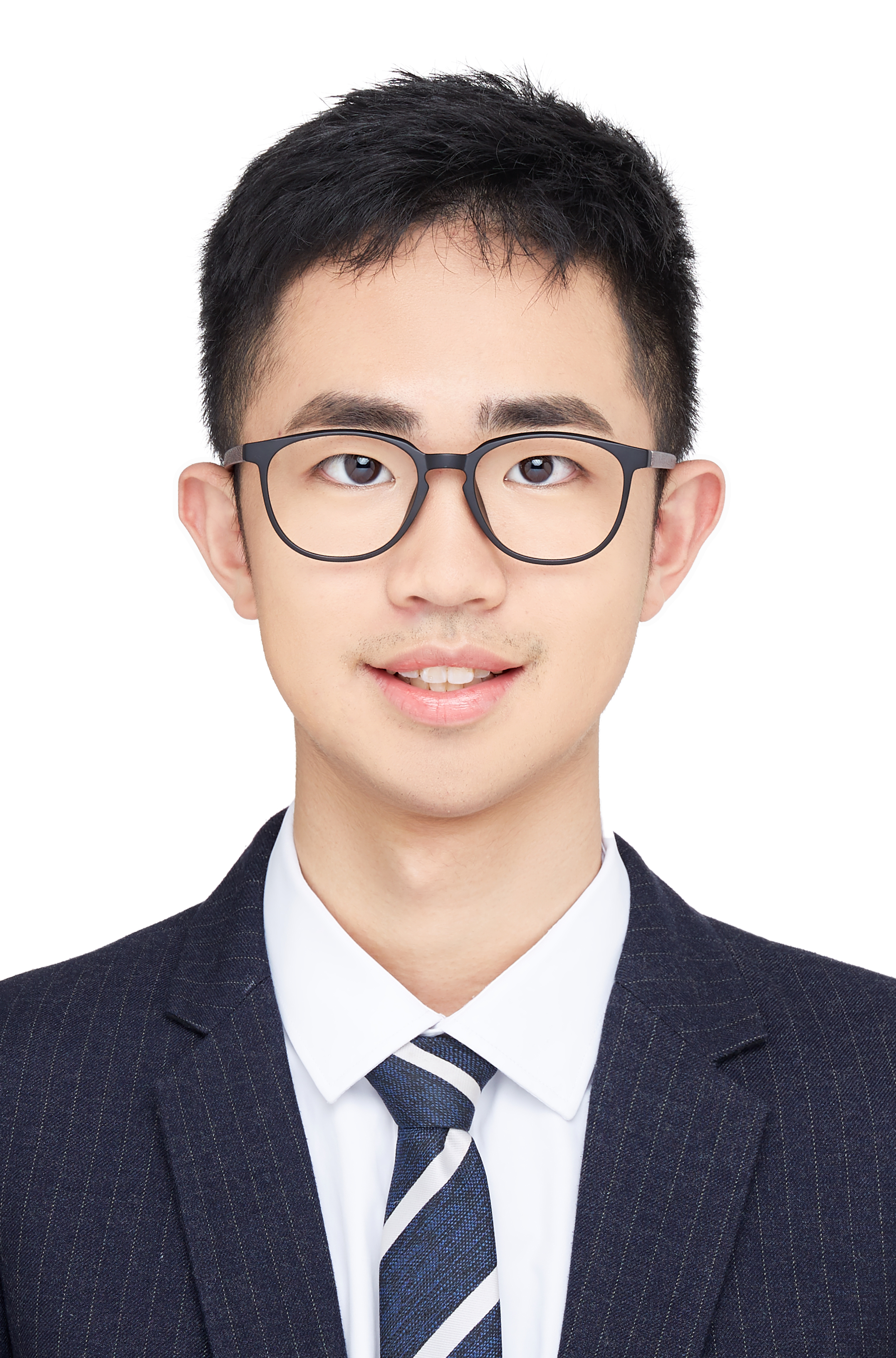}}]{Chunyuan Zheng} is currently a Ph.D. student at the School of Mathematical Sciences, Peking University. He received his M.S. degree in statistics from the University of California, San Diego, in 2024. He has published over 20 papers in leading conferences and journals, including NeurIPS, ICML, ICLR, KDD, and TOIS. He serves as an Area Chair for NeurIPS 2026 and has co-chaired workshops at ICLR, AAAI, and ICDM. His research interests include causal inference, trustworthy machine learning, and large language model post-training.
\end{IEEEbiography}

\begin{IEEEbiography}[{\includegraphics[width=1in,height=1.25in,clip,keepaspectratio]{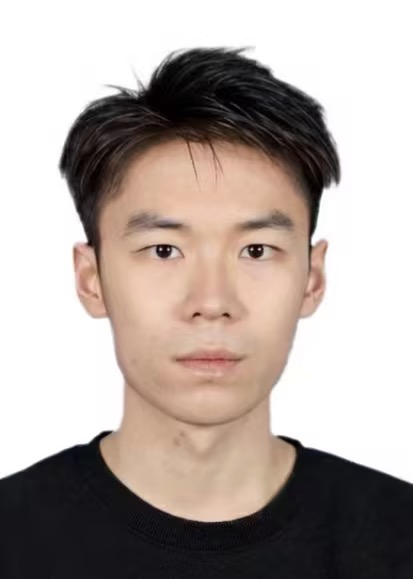}}]{Xiaoxi Li} is currently a PhD student at the Gaoling School of Artificial Intelligence, Renmin University of China.
He earned his bachelor's degree at Nankai University in 2023.
As the first author, he has published several papers in top-tier AI conferences and journals such as TOIS, SIGIR, ACL, EMNLP, etc.
He is currently working on sequential modeling, retrieval-augmented generation, large language models and deep research agents.
\end{IEEEbiography}

\begin{IEEEbiography}[{\includegraphics[width=1in,height=1.25in,clip,keepaspectratio]{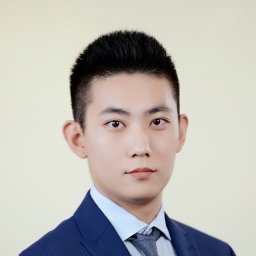}}]{Zhixuan Chu} is an associate professor with the College of Computer Science and Engineering, Zhejiang University.
He has published more than 40 papers in top-tier AI conferences such as ICML, NeurIPS, ICLR.
He has regularly been served as the PC member or Area Chair for top conferences such as ICML, NeurIPS, ICLR.
His research interests include causal inference, security, and time-series modeling.
\end{IEEEbiography}

\begin{IEEEbiography}[{\includegraphics[width=1in,height=1.25in,clip,keepaspectratio]{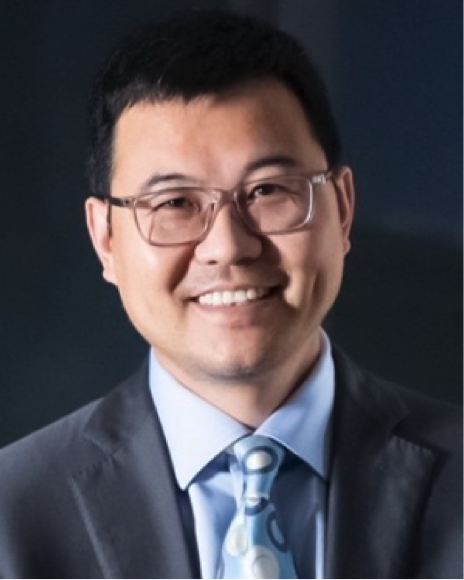}}]{Chao Xu} (Senior Member, IEEE) is a full professor with the College of Control Science and Engineering, Zhejiang University.
He is also the Inaugural Dean of the ZJU Huzhou Institute.
He has published over 100 articles in international journals, including Science Robotics and Nature Machine Intelligence.
His research interests include geometric learning \& analytical dynamics, with applications in control systems and robotics.
He served as the organization committee of IROS-2025.
\end{IEEEbiography}

\begin{IEEEbiography}[{\includegraphics[width=1in,height=1.25in,clip,keepaspectratio]{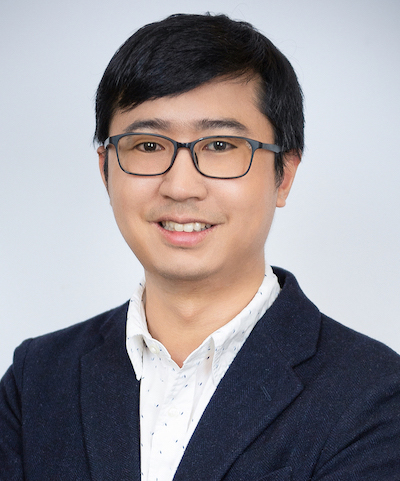}}]{Mingming Gong}
is an associate professor at University of Melbourne and an affiliated associate professor at the Mohamed bin Zayed University of Artificial Intelligence.
He is interested in learning the causal generative processes behind data and is also drawn to applied machine learning areas such as 3D vision, LLMs, and VLMs.
He has authored more than 100 papers in top venues such as ICML, NeurIPS, ICLR, JMLR, TPAMI, and served as a meta-reviewer for NeurIPS, ICML, ICLR, MLJ, TMLR, etc.
He is a recepient of the Australian Artificial Intelligence Emerging Researcher Award and the Australian Research Council Early Career Researcher Award.
\end{IEEEbiography}

\begin{IEEEbiography}[{\includegraphics[width=1in,height=1.25in,clip,keepaspectratio]{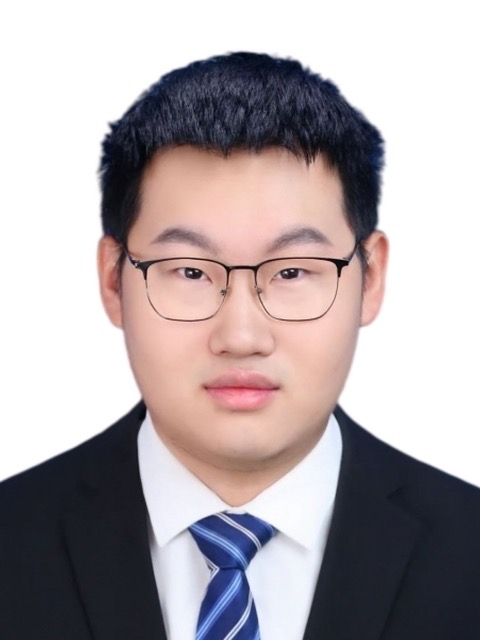}}]{Haoxuan Li} is an assistant professor in the Institute for Artificial Intelligence at Peking University.
He has more than 80 publications appeared in top-tier AI conferences such as ICML, NeurIPS, ICLR, SIGKDD, WWW, SIGIR, CVPR, ICDE, and ACL, reported by MIT Technology Review.
He has served as the area chair (AC) for top-tier conferences including ICML, NeurIPS, ICLR, SIGKDD, and the invited reviewer for prestigious journals such as TPAMI, TKDE, TOIS, TKDD, and TNNLS.
\end{IEEEbiography}

\begin{IEEEbiography}[{\includegraphics[width=1in,height=1.25in,clip,keepaspectratio]{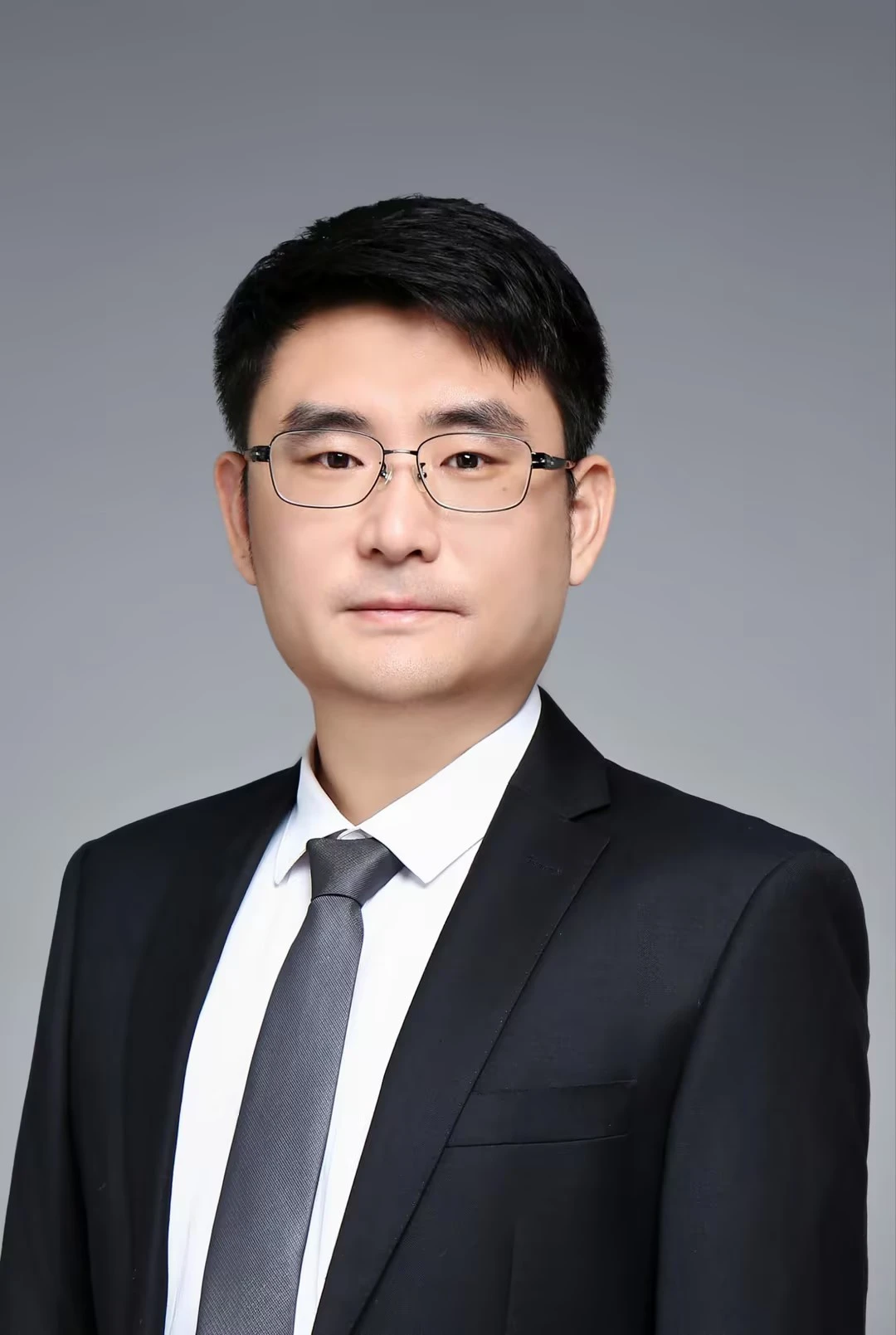}}]{Yuan Lu}
    is currently the technical lead of the Platform Algorithm Team at Xiaohongshu.
Prior to joining Xiaohongshu, he worked at DiDi AI Lab and Alibaba DAMO Academy, where he focused on research and development in pre-trained models for over 7 years.
His research interests include representation learning, dialogue systems, graph neural networks, and pre-training for both language and multimodal models.
\end{IEEEbiography}

\begin{IEEEbiography}[{\includegraphics[width=1in,height=1.25in,clip,keepaspectratio]{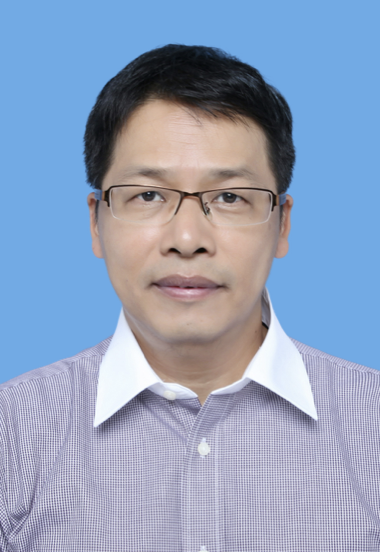}}]{Zhouchen Lin} (Fellow, IEEE)
received the Ph.D. degree in applied mathematics from Peking University in 2000.
He is currently a Boya Special Professor with the State Key Laboratory of General Artificial Intelligence, School of Intelligence Science and Technology, Peking University.
His research interests include machine learning and numerical optimization.
He has published over 310 papers, collecting more than 40,000 Google Scholar citations.
He is a Fellow of the IAPR, the IEEE, the AAIA and the CSIG.
He also regularly serves as the Senior Area Chair for top conferences including NeurIPS, ICML, KDD, etc, the Board Member of ICML, and the Associate Editor-in-Chief for IEEE TPAMI.
\end{IEEEbiography}

\begin{IEEEbiography}[{\includegraphics[width=1in,height=1.25in,clip,keepaspectratio]{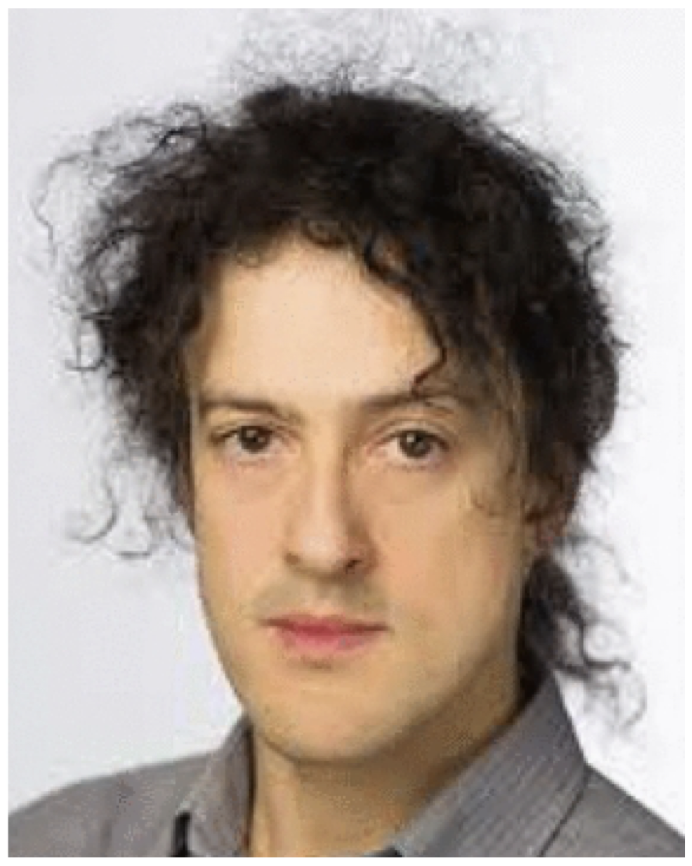}}]{Philip Torr} 
    (Senior Member, IEEE) is a professor in the Department of Engineering Science at University of Oxford.
He has won awards from top vision conferences, including ICCV, CVPR, ECCV, NIPS, and BMVC, collecting more than 140,000 Google Scholar citations.
He is a Fellow of the Royal Academy of Engineering (FREng), Fellow of the Royal Society (FRS), Turing AI world leading researcher fellow in 2021, and Schmidt Sciences AI2050 Senior Fellowship in 2025.
\end{IEEEbiography}

\begin{IEEEbiography}[{\includegraphics[width=1in,height=1.25in,clip,keepaspectratio]{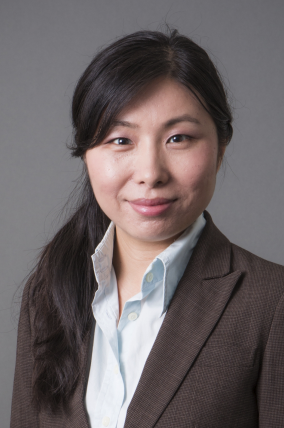}}]{Yan Liu} 
    (Fellow, IEEE) is a full professor in the Computer Science Department, Viterbi School of Engineering at University of Southern California.
She was an assistant professor from 2010 to 2016, and an associate professor from 2016 to 2020.
Before joining USC, she was a research staff member at the IBM T.J. Watson Research Center from 2006 to 2010.
She received her M.S. and Ph.D. from Carnegie Mellon University.
Her research interests include machine learning with applications to health, sustainability, and social media.
She served as general chair  for ICLR 2023 and ACM KDD 2020, also as the associate Editor-in-Chief of TPAMI and Board Member and Finance Chair of ICLR. 
\end{IEEEbiography}

\clearpage
\onecolumn
\normalsize
\appendix

\section{Empirical Results for Different Losses}

In this section, we compare the empirical performance of different loss functions. We use the FredFormer~\citep{fredformer} as the backbone model, and evaluate the performance on the ETTm1, ETTh1~\citep{Informer}, ECL, and Weather~\citep{Autoformer} datasets. The results are reported in \autoref{tab:loss-app} with observations as follows.
\ding{182} \textbf{Loss functions that account for label autocorrelation generally outperform the pointwise MSE loss.} QDF, DistDF, Time-o1, FreDF, and DBLoss all achieve lower average MSE and MAE on all four datasets. For example, on ECL, DistDF reduces the average MSE from 0.191 to 0.173, while DBLoss reduces the average MAE from 0.284 to 0.264. These results suggest that accounting for label autocorrelation during training can improve forecasting accuracy beyond minimizing pointwise errors alone.
\ding{183} \textbf{No single loss function uniformly dominates across datasets, horizons, and metrics.} QDF and DistDF achieve the lowest average MSE on the two ETT datasets, DistDF attains the best average MSE on ECL, whereas DBLoss obtains the lowest average MAE on ETTm1, ECL, and Weather. This variation indicates that the relative benefit of a loss depends on the data characteristics.

\begin{table}
\caption{Comparable results with different loss functions. }\label{tab:loss-app}
\renewcommand{\arraystretch}{0.9}
\setlength{\tabcolsep}{5pt}
\scriptsize
\centering
\renewcommand{\multirowsetup}{\centering}
\begin{threeparttable}
\begin{tabular}{c|c|cc|cc|cc|cc|cc|cc|cc|cc|cc}
\toprule
\multicolumn{2}{l}{\multirow{2}{*}{\rotatebox{0}{Loss}}} & 
\multicolumn{2}{c}{\rotatebox{0}{QDF}} &
\multicolumn{2}{c}{\rotatebox{0}{DistDF}} &
\multicolumn{2}{c}{\rotatebox{0}{Time-o1}} &
\multicolumn{2}{c}{\rotatebox{0}{FreDF}} &
\multicolumn{2}{c}{\rotatebox{0}{DBLoss}} &
\multicolumn{2}{c}{\rotatebox{0}{Dilate}} &
\multicolumn{2}{c}{\rotatebox{0}{Soft-DTW}} &
\multicolumn{2}{c}{\rotatebox{0}{DTW}} &
\multicolumn{2}{c}{\rotatebox{0}{MSE}} \\
\multicolumn{2}{c}{} &
\multicolumn{2}{c}{(2026)} &
\multicolumn{2}{c}{(2026)} & 
\multicolumn{2}{c}{(2026)} & 
\multicolumn{2}{c}{(2025)} & 
\multicolumn{2}{c}{(2025)} & 
\multicolumn{2}{c}{(2019)} & 
\multicolumn{2}{c}{(2017)} & 
\multicolumn{2}{c}{(2003)} & 
\multicolumn{2}{c}{(-)} \\
\cmidrule(lr){3-4} \cmidrule(lr){5-6}\cmidrule(lr){7-8} \cmidrule(lr){9-10}\cmidrule(lr){11-12} \cmidrule(lr){13-14} \cmidrule(lr){15-16} \cmidrule(lr){17-18} \cmidrule(lr){19-20}
\multicolumn{2}{l}{\rotatebox{0}{Metrics}}  & MSE & MAE  & MSE & MAE  & MSE & MAE  & MSE & MAE  & MSE & MAE  & MSE & MAE  & MSE & MAE  & MSE & MAE  & MSE & MAE \\
\toprule

\multirow{5}{*}{{\rotatebox{90}{\scalebox{0.95}{ETTm1}}}}
& 96 & 0.312 & 0.353 & 0.316 & 0.357 & 0.321 & 0.357 & 0.326 & 0.355 & 0.317 & 0.346 & 0.337 & 0.367 & 0.332 & 0.363 & 0.332 & 0.364 & 0.326 & 0.361 \\
& 192 & 0.359 & 0.380 & 0.358 & 0.380 & 0.360 & 0.378 & 0.363 & 0.380 & 0.365 & 0.374 & 0.364 & 0.384 & 0.370 & 0.386 & 0.370 & 0.386 & 0.365 & 0.382 \\
& 336 & 0.391 & 0.403 & 0.392 & 0.404 & 0.389 & 0.400 & 0.392 & 0.400 & 0.395 & 0.398 & 0.397 & 0.406 & 0.406 & 0.409 & 0.409 & 0.410 & 0.396 & 0.404 \\
& 720 & 0.450 & 0.438 & 0.448 & 0.437 & 0.447 & 0.435 & 0.454 & 0.440 & 0.454 & 0.433 & 0.457 & 0.443 & 0.478 & 0.450 & 0.476 & 0.448 & 0.459 & 0.444 \\
\cmidrule(lr){2-20}
& Avg & 0.378 & 0.393 & 0.378 & 0.394 & 0.379 & 0.393 & 0.384 & 0.393 & 0.383 & 0.388 & 0.389 & 0.400 & 0.397 & 0.402 & 0.396 & 0.402 & 0.387 & 0.398 \\
\midrule

\multirow{5}{*}{{\rotatebox{90}{\scalebox{0.95}{ETTh1}}}}
& 96 & 0.367 & 0.392 & 0.373 & 0.393 & 0.368 & 0.391 & 0.370 & 0.392 & 0.368 & 0.389 & 0.378 & 0.399 & 0.376 & 0.398 & 0.378 & 0.399 & 0.377 & 0.396 \\
& 192 & 0.421 & 0.425 & 0.428 & 0.425 & 0.424 & 0.422 & 0.426 & 0.421 & 0.432 & 0.424 & 0.439 & 0.435 & 0.439 & 0.435 & 0.438 & 0.433 & 0.437 & 0.425 \\
& 336 & 0.468 & 0.447 & 0.466 & 0.445 & 0.467 & 0.441 & 0.467 & 0.439 & 0.455 & 0.442 & 0.481 & 0.453 & 0.484 & 0.455 & 0.486 & 0.455 & 0.486 & 0.449 \\
& 720 & 0.464 & 0.473 & 0.453 & 0.453 & 0.465 & 0.463 & 0.474 & 0.466 & 0.477 & 0.460 & 0.516 & 0.482 & 0.542 & 0.510 & 0.538 & 0.510 & 0.488 & 0.467 \\
\cmidrule(lr){2-20}
& Avg & 0.430 & 0.434 & 0.430 & 0.429 & 0.431 & 0.429 & 0.434 & 0.429 & 0.433 & 0.429 & 0.453 & 0.442 & 0.460 & 0.449 & 0.460 & 0.449 & 0.447 & 0.434 \\
\midrule

\multirow{5}{*}{{\rotatebox{90}{\scalebox{0.95}{ECL}}}}
& 96 & 0.153 & 0.244 & 0.145 & 0.238 & 0.151 & 0.245 & 0.149 & 0.240 & 0.154 & 0.241 & 0.158 & 0.253 & 0.168 & 0.266 & 0.158 & 0.253 & 0.161 & 0.258 \\
& 192 & 0.164 & 0.254 & 0.162 & 0.255 & 0.166 & 0.256 & 0.165 & 0.255 & 0.165 & 0.250 & 0.170 & 0.263 & 0.218 & 0.313 & 0.216 & 0.307 & 0.174 & 0.269 \\
& 336 & 0.180 & 0.271 & 0.176 & 0.270 & 0.181 & 0.274 & 0.178 & 0.271 & 0.181 & 0.269 & 0.190 & 0.286 & 0.197 & 0.291 & 0.199 & 0.295 & 0.194 & 0.290 \\
& 720 & 0.210 & 0.297 & 0.211 & 0.300 & 0.213 & 0.304 & 0.213 & 0.300 & 0.214 & 0.296 & 0.229 & 0.316 & 0.240 & 0.322 & 0.235 & 0.322 & 0.235 & 0.319 \\
\cmidrule(lr){2-20}
& Avg & 0.177 & 0.266 & 0.173 & 0.266 & 0.178 & 0.270 & 0.176 & 0.266 & 0.178 & 0.264 & 0.187 & 0.280 & 0.206 & 0.298 & 0.202 & 0.294 & 0.191 & 0.284 \\
\midrule

\multirow{5}{*}{{\rotatebox{90}{\scalebox{0.95}{Weather}}}}
& 96 & 0.171 & 0.214 & 0.172 & 0.212 & 0.171 & 0.208 & 0.171 & 0.210 & 0.174 & 0.209 & 0.173 & 0.214 & 0.173 & 0.213 & 0.179 & 0.219 & 0.180 & 0.220 \\
& 192 & 0.219 & 0.254 & 0.218 & 0.255 & 0.219 & 0.253 & 0.217 & 0.252 & 0.219 & 0.249 & 0.225 & 0.260 & 0.220 & 0.255 & 0.223 & 0.257 & 0.222 & 0.258 \\
& 336 & 0.277 & 0.296 & 0.277 & 0.297 & 0.277 & 0.295 & 0.275 & 0.295 & 0.278 & 0.293 & 0.280 & 0.299 & 0.281 & 0.296 & 0.281 & 0.298 & 0.283 & 0.301 \\
& 720 & 0.353 & 0.346 & 0.352 & 0.347 & 0.353 & 0.346 & 0.352 & 0.346 & 0.355 & 0.345 & 0.355 & 0.348 & 0.369 & 0.355 & 0.356 & 0.347 & 0.358 & 0.348 \\
\cmidrule(lr){2-20}
& Avg & 0.255 & 0.278 & 0.255 & 0.277 & 0.255 & 0.276 & 0.254 & 0.276 & 0.256 & 0.274 & 0.258 & 0.280 & 0.261 & 0.280 & 0.260 & 0.280 & 0.261 & 0.282 \\

\bottomrule
\end{tabular}
\end{threeparttable}
\end{table}

\end{document}